\documentclass[10pt,twocolumn,letterpaper]{article}

\usepackage{iccv}      

\definecolor{iccvblue}{rgb}{0.21,0.49,0.74}
\usepackage[pagebackref,breaklinks,colorlinks,allcolors=iccvblue]{hyperref}
\usepackage[accsupp]{axessibility}  

\usepackage[utf8]{inputenc}
\usepackage{graphicx}
\usepackage{adjustbox}
\usepackage{pifont}
\usepackage{tikz}
\usepackage{amssymb}
\usepackage{relsize}
\newcommand{\tikzxmark}{%
\tikz[scale=0.15] {
    \draw[line width=0.7,line cap=round] (0,0) to [bend left=6] (1,1);
    \draw[line width=0.7,line cap=round] (0.2,0.95) to [bend right=3] (0.8,0.05);
}}
\newcommand{\tikzcmark}{%
\tikz[scale=0.18] {
    \draw[line width=0.7,line cap=round] (0.25,0) to [bend left=10] (1,1);
    \draw[line width=0.8,line cap=round] (0,0.35) to [bend right=1] (0.23,0);
}}

\usepackage{xr} 
\usepackage{multirow}
\usepackage{arydshln}

\def\paperID{13789} 
\def\confName{ICCV}
\def\confYear{2025}

\title{Joint Learning of Pose Regression and Denoising Diffusion with Score Scaling Sampling for Category-level 6D Pose Estimation}

\author{Seunghyun Lee \quad Tae-Kyun Kim \\
KAIST\\[-0.1em]
  {\tt\small \{sseunghyun, kimtaekyun\}@kaist.ac.kr} \vspace{-0.7em}
}

\begin{document}
\maketitle
\begin{abstract}
Latest diffusion models have shown promising results in category-level 6D object pose estimation by modeling the conditional pose distribution with depth image input. The existing methods, however, suffer from slow convergence during training, learning its encoder with the diffusion denoising network in end-to-end fashion, and require an additional network that evaluates sampled pose hypotheses to filter out low-quality pose candidates. In this paper, we propose a novel pipeline that tackles these limitations by two key components. First, the proposed method pretrains the encoder with the direct pose regression head, and jointly learns the networks via the regression head and the denoising diffusion head, significantly accelerating training convergence while achieving higher accuracy. Second, sampling guidance via time-dependent score scaling is proposed s.t. the exploration-exploitation trade-off is effectively taken, eliminating the need for the additional evaluation network. The sampling guidance maintains multi-modal characteristics of symmetric objects at early denoising steps while ensuring high-quality pose generation at final steps. Extensive experiments on multiple benchmarks including REAL275, HouseCat6D, and ROPE, demonstrate that the proposed method, simple yet effective, achieves state-of-the-art accuracies even with single-pose inference, while being more efficient in both training and inference.
\end{abstract}    

\section{Introduction}
\label{sec:intro}

6D object pose estimation aims to predict the 3D rotation and translation of objects from RGB or depth images. It is crucial for a wide range of real-world applications, including robotics \cite{kappler2018real,mousavian20196,deng2020self,tremblay2018deep, hodan2018bop}, augmented reality \cite{nee2012augmented, marchand2015pose, tjaden2017real}, and 3D scene understanding \cite{nie2020total3dunderstanding, zhang2021holistic, cho2024dense}. While previous methods achieved promising results in instance-level object pose estimation \cite{labbe2020cosypose, wang2019densefusion, peng2019pvnet, sock2020introducing, castro2023crt}, their reliance on CAD models of target object instances limits its applications in practical scenarios. On the other hand, category-level pose estimation \cite{sahin2018category, wang2019normalized} has emerged as a valuable alternative, since it does not require instance-level CAD models. This approach has gained considerable attention with its capability to handle novel instances within known categories, making it highly suitable for real-world situations.

Recent advances in diffusion models have brought notable progresses to category-level 6D pose estimation \cite{zhang2023genpose, zhang2025omni6dpose, ikeda2024diffusionnocs}. GenPose \cite{zhang2023genpose} reframed conventional regression-based pose estimation as a generative task, effectively addressing the multi-hypothesis issue—a challenging scenario where multiple pose candidates can be equally valid due to object symmetries or partial observations \cite{manhardt2019explaining}. With an encoder network of depth images as partial point clouds and a pose diffusion model conditioned on the encoder representation, 
it effectively captures the underlying pose distribution without requiring any symmetry-aware designs~\cite{lin2022sar, di2022gpv} or shape priors~\cite{lin2021dualposenet, lin2022category, zhang2022rbp} that were common in previous approaches. This pipeline is shown to be effective by handling pose ambiguity arising from shape variations and symmetric objects within categories.


Despite its SOTA performance, diffusion-based pose estimation framework exhibits a significant bottleneck: slow training convergence. This limitation extends beyond GenPose to various approaches that model different pose representation spaces through diffusion processes~\cite{wang2023posediffusion, xu20246d, wang2025text}. The challenge primarily stems from the necessity of tracking extensive forward and reverse diffusion trajectories across multiple time steps~\cite{hang2023efficient, zhang2024improving}, while the encoder network is learnt at various noise levels in an end-to-end fashion. In contrast to large volume of work based on Latent Diffusion Models (LDMs) which benefit from pre-trained and fixed encoders~\cite{rombach2022high, gu2022vector, blattmann2023align, vahdat2022lion, esser2023structure}, pose diffusion models typically optimize both feature extraction and denoising processes concurrently. This learning process through score-matching objectives substantially increases computational complexity and learning epochs.

Another challenge in diffusion-based pose estimation frameworks lies in their tendency to generate outliers that deviate from the conditional distribution during the stochastic sampling process. When generating multiple pose hypotheses, these frameworks inevitably produce some outlier poses that do not align with the true distribution. How to effectively filter these outliers and aggregate the remaining valid poses into a single, accurate estimate remains an open research question across various works~\cite{shan2023diffusion, holmquist2023diffpose, zhang2023genpose, zhang2025omni6dpose}. In the context of 6D pose estimation, GenPose addresses this limitation by adopting an additional network to rank and filter out these low-density sampled poses. However, the reliance on an extra network beyond the pose generation model increases the overall pipeline complexity. Furthermore, using mean-pooling to aggregate the pose candidates can lead to mode-averaging issues, especially for objects with discrete symmetries.

In this paper, we introduce two key components that address the aforementioned limitations. First, we pre-train the encoder with direct pose regression objectives before jointly training it with the diffusion head. 
This provides complementary supervision signals—the score matching loss guides the encoder to learn consistent features under noise, while the regression loss offers direct guidance for precise pose estimation. The combined effect of these losses enables the encoder to learn robust and accurate feature representations, significantly accelerating convergence while improving the final accuracy. Second, a time-dependent score-scaling diffusion sampling is adopted, eliminating the need for post-processing outlier removal. Simply scaling the score in a time-dependent manner, the diffusion samples are strongly guided toward high-density regions under the partial point cloud condition. This simple yet effective approach enables reliable pose estimation even with a single pose sampling, thus avoiding both the computational overhead of additional likelihood estimation networks and the mean-mode issues that arise during multi-pose aggregation.

In summary, our key contributions are as follows:
\begin{itemize}
    \item We propose a novel training strategy combining direct pose regression and diffusion score-matching objectives. By leveraging complementary supervision signals, this strategy significantly accelerates training convergence while improving pose estimation accuracy.
    
    \item We introduce a time-dependent score-scaling diffusion sampling method that effectively guides samples toward high-density regions, enabling reliable single-sample pose estimation without requiring additional networks or complex post-processing steps.
    \item Through extensive experiments on multiple benchmarks, we demonstrate that our method achieves state-of-the-art performance while significantly reducing training time and computational overhead compared to previous relevant approaches.
\end{itemize}

\begin{figure*}[h!]
     \centering
     \includegraphics[width=0.8\textwidth]{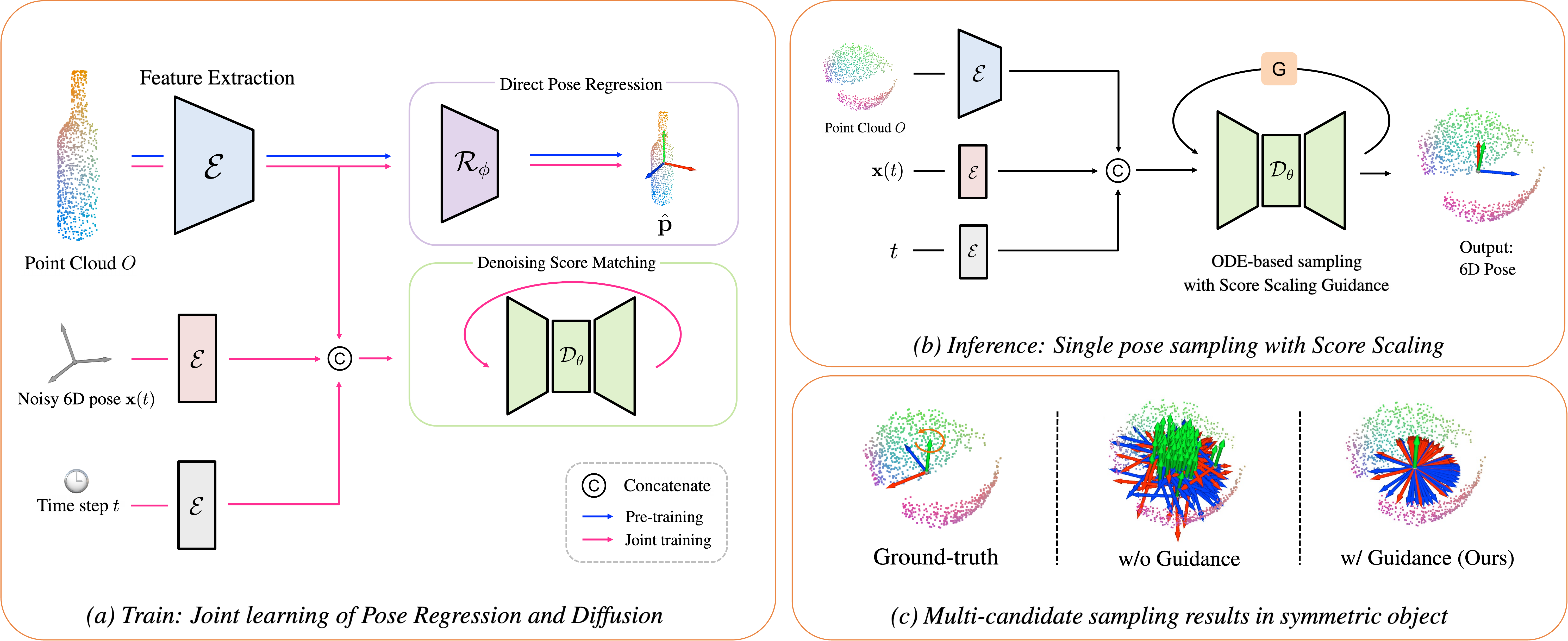}
     \caption{\textbf{(a) Joint learning phase}: The encoder $\mathcal{E}$ and regression head $\mathcal{R}_{\phi}$ are first pre-trained on target data, then jointly trained with the diffusion head $\mathcal{D}_{\theta}$. \textbf{(b) Inference phase}: Single pose sampling using score scaling guidance $w_t$ to update scores at each timestep. \textbf{(c) Sampling results on symmetric objects}: showing that our score scaling guidance prevents outlier poses while preserving the symmetric distribution, compared to sampling without guidance.}
     \label{fig:overview}
     \vspace{-0.5cm}
 \end{figure*}

\section{Related Works}
\label{sec:related}


\textbf{Regression-based approaches.} Category-level 6D pose estimation~\cite{sahin2018category, chen2021fs, chen2021sgpa, liu2023net, lin2022category, wang2019normalized} addresses the challenging task of predicting poses for unseen instances within known categories. Most existing methods rely on regression-based approaches to tackle this problem. NOCS~\cite{wang2019normalized} introduced a canonical representation called normalized object coordinate space (NOCS), shared across object instances within the category. Object pose is estimated by regressing the NOCS coordinates of observed points and then applying the pose alignment algorithm. SPD~\cite{tian2020shape} and CR-Net~\cite{wang2021category} enhanced this regression framework by learning the deformation of category-level shape priors to reconstruct an accurate canonical model. Another line of regression-based methods directly estimates pose in an end-to-end manner: DualPoseNet~\cite{lin2021dualposenet}  simultaneously regresses the pose parameters and reconstructs the point cloud in its canonical pose, while SSP-pose~\cite{zhang2022ssp} incorporates shape priors into a direct pose regression network. Despite promising results, regression-based approaches are limited by single-pose supervision, making it difficult to handle multiple valid pose hypotheses in symmetry or partial observations.\\
\noindent \textbf{Generative approaches.} Recent studies have reformulated category-level pose estimation as a generative modeling task to address the aforementioned pose ambiguity problem. DiffusionNOCS~\cite{ikeda2024diffusionnocs} leverages diffusion models to estimate dense canonical maps from multi-modal image inputs, facilitating object pose and partial geometry recovery. GenPose~\cite{zhang2023genpose} and GenPose++~\cite{zhang2025omni6dpose} directly model the conditional pose distribution using score-based diffusion models, with GenPose++ further incorporating semantic RGB features along with depth image input. Both methods employ an additional energy-based network to filter low-quality pose samples by estimating their likelihood. Despite its state-of-the-art performance, GenPose faces slow training convergence due to end-to-end learning with the diffusion network and the encoder, and requires an additional network for pose candidate evaluation, which increases complexity and computational overhead. Moreover, existing approaches have yet to explore the potential integration of direct pose regression with diffusion-based generation, indicating opportunities to enhance both training efficiency and accuracy.\\
\noindent \textbf{Guidance in Diffusion Models.} Guidance in diffusion models has become essential for enhancing generation quality, with Classifier-Free Guidance (CFG)~\cite{ho2021classifierfree} being the most widely adopted approach. CFG guides the diffusion sampling process by extrapolating between the conditional and unconditional denoising results, effectively increasing the conditional probability to improve fidelity at the cost of diversity. 
Recent works have improved upon this by introducing dynamic guidance strategies. For example, linearly increasing the guidance weight during sampling potentially increases diversity~\cite{chang2023muse, blattmann2023stable}. Another approach involves applying guidance selectively within specific noise intervals~\cite{Kynkaanniemi2024INT}. These strategies respond to observations that CFG influences noise levels differently—having negative effects in early (high-noise) steps but beneficial impacts in middle steps.
Moving beyond the conditional-unconditional paradigm, Karras \textit{et al.}~\cite{Karras2024AUTO} propose using an inferior version of the conditional model as guidance, showing that guiding away from an under-trained yet compatible model can improve sample quality while preserving diversity compared to traditional CFG. However, existing methods mostly follow the CFG framework of combining two denoising results, whereas our approach scales the output score directly. Although naive score lengthening has previously been observed to cause over-smoothing and reduced image diversity~\cite{Karras2024AUTO}, we demonstrate that, with appropriate scaling scheduling, this straightforward yet effective strategy not only improves pose accuracy but also preserves the distribution of multiple valid poses that can occur with symmetric and partially observed objects.\\
\noindent \textbf{Pre-training 3D Point cloud Encoders.}
Recent 3D point cloud understanding has advanced through supervised approaches using datasets such as ShapeNet~\cite{chang2015shapenet}, ShapeNetPart~\cite{yi2016scalable} and ModelNet40~\cite{wu20153d} as well as self-supervised methods. Wang \textit{et al.}~\cite{wang2021unsupervised} and Xie \textit{et al.}~\cite{xie2020pointcontrast} have implemented self-supervised learning by contrasting different views and completing occluded regions. Pang \textit{et al.}~\cite{pang2022masked} and Yu \textit{et al.}~\cite{yu2022point} further demonstrated effective representation learning by adapting masked autoencoders to point clouds. However, these pre-training approaches have shown limited transferability to 6D pose estimation, where precise object localization requires specialized techniques beyond general representation learning.

\section{Preliminary}
\label{sec:preliminary}

\paragraph{Denoising Diffusion.} Denoising diffusion models~\cite{song2019generative, song2020score, ho2020denoising} is a class of generative models that can transform noise into samples from the target data distribution through an iterative denoising process. The key idea is to gradually corrupt data samples with Gaussian noise through a forward process and then learn the model to reverse this corruption.
Formally, let  {\small${p(\mathbf{x}(t))}$} denote a perturbed data distribution obtained by convolving the data distribution {\small$\smash{p_{data}(\mathbf{x})}$} with a Gaussian kernel {\small$\smash{\mathcal{N}(\mathbf{x}; \mathbf{0}, \sigma_t^2\mathbf{I})}$}, where $\sigma_t$ represents the noise level at time $t$.  For a large enough $\sigma_T$,  {\small$\smash{p(\mathbf{x}(T))}$} closely approximates {\small$\smash{\mathcal{N}(\mathbf{x}; 0, \sigma_{T}^2 I)}$}, from which we can easily sample a data by drawing from a gaussian noise. The sample is then gradually transformed to lower noise levels by following a probability flow ODE~\cite{song2020score, karras2022elucidating} :
\begin{equation} \label{eq:1}
d\mathbf{x} =  -\sigma_t \dot \sigma_t\nabla_{\mathbf{x}}\log p(\mathbf{x}(t)){dt} \end{equation} 
where the dot represents a time derivative. In practice, Equation \ref{eq:1} is solved numerically by taking steps along the trajectory~\cite{song2020score}. The score function {\small$\smash{\nabla_{\mathbf{x}}\log p(\mathbf{x};\sigma_t)}$} is a vector field pointing toward regions of higher data density at a given noise level~\cite{karras2022elucidating}, and it can be approximated by a neural network {\small$\smash{\mathbf{s}_{\theta}(\mathbf{x}(t), t)}$} trained via the Denoising Score Matching objective~\cite{vincent2011connection}:
\begin{small}
\begin{equation} \label{eq:2}
\mathcal{L}(\theta) =  \mathbb{E}_{t}\bigg\{\lambda(t)\mathbb{E}_{\mathbf{x}(0), \\ {\mathbf{x}(t)}}\bigg[ \bigg\| \mathbf{s}_\theta(\mathbf{x}(t),t)  + \frac{\mathbf{x}(t)-\mathbf{x}(0)}{\sigma_t^2}\bigg\|_2^2\bigg]\bigg\} 
\end{equation}\end{small}
Here, {\small$\smash{\lambda}$} is a weighting function, {\small$\smash{t ~\sim \mathcal{U}(0,T)}$}, {\small$\smash{\mathbf{x}(0)}$}{\small$\sim p_{data}(\mathbf{x})$}  and {\small$\smash{\mathbf{x}(t)\sim \mathcal{N}(\mathbf{x}(t);\mathbf{x}(0),\sigma_t^2 \mathbf{I})}$}. Minimizing the objective in Equation \ref{eq:2} ensures that the optimal score network satisfies {\small$\smash{\mathbf{s}_\theta(\mathbf{x}(t), t)=\nabla_{\mathbf{x}} \log p(\mathbf{x}(t))}$} , as shown in ~\cite{vincent2011connection, karras2022elucidating}.

\paragraph{GenPose.} GenPose is the first framework to successfully adapt the denoising diffusion process for category-level 6D pose estimation using depth images. Given a partially observed point cloud $O$, it aims to model the conditional distribution {\small$\smash{p(\mathbf{p} | O)}$} of object poses, where {\small$\smash{\mathbf{p} \in SE(3)}$} represents the 6D pose of the object.
To achieve this, GenPose first extracts a global feature of the point cloud using a PointNet++~\cite{qi2017pointnet++} encoder. Combining this feature with the noisy pose and time step $t$ as input, the diffusion model learns a score function following Equation \ref{eq:2}. The encoder is not pre-trained and learnt with the denoising model in end-to-end, taking slow training convergence. Once the model converges, 6D pose can be sampled from pure white noise by solving Equation ~\ref{eq:1}.
In GenPose, K pose candidates are sampled given the same observed point cloud condition. However, some poses may be outliers with low likelihood, falling outside the expected distribution. To address this, GenPose employs an energy-based model, another diffusion model, to estimate the likelihood of each pose candidate. Low-scoring candidates are filtered out based on these energy scores, and the top $\smash{\delta\%}$ of remaining poses are then aggregated through mean-pooling to produce the final pose.

\section{Proposed method}
\label{sec:method}
\vspace{-0.2em} 
\subsection{Overview}
The overall framework of our proposed approach is illustrated in Figure \ref{fig:overview}. Following GenPose; we consider the category level 6D pose estimation task, where the goal is to estimate the SE(3) pose of arbitrary object instances within the same category from partial observations. Specifically, given a point cloud {\small$\smash{O \in \mathbb{R}^{1024 \times 3}}$} obtained by back-projecting the segmented depth data, we aim to model the conditional pose distribution {\small$\smash{p(\mathbf{p}|O)}$}. Our framework first extracts point cloud features using PointNet++~\cite{qi2017pointnet++}. The noisy pose {\small$\smash{\mathbf{p}(t)}$} and time step $t$ are also embedded through MLPs. These embedded values are then combined with the point cloud features and fed into the denoising diffusion head $\smash{\mathcal{D}_{\theta}}$ to learn the score function. Simultaneously, the same point cloud features are passed to a regression head $\smash{R_{\phi}}$, which directly computes the loss with ground truth poses. These two heads are trained end-to-end in a combined manner. During inference, we utilize only the  diffusion head. Starting from a single initial noise, we estimate the score function at each time step t to sample the 6D pose. By appropriately scaling the magnitude of the score at each time step, we can effectively push the samples toward high-density regions of the conditional distribution, preventing the generation of out-of-distribution poses without the need of an additional pose evaluation network. 
Later in the experiments (Section \ref{sec:exp}), we show that this simple joint learning strategy accelerates model convergence and enhances performance. Furthermore, our score scaling approach achieves superior performance with a single guided pose compared to previous methods that require multiple pose sampling and aggregation.

\subsection{Joint learning phase}
We propose a two-stage learning strategy that combines pose regression and denoising diffusion objectives to leverage both direct pose supervision and multi-modal pose distribution learning. The learning process consists of pre-training and joint training stages.

\paragraph{Pre-training.} We first pre-train the encoder with direct pose regression to establish a strong prior for capturing 6D pose information from point clouds (see the Supplementary for more discussions). The encoder extracts a 1024-dimensional feature vector from the input point cloud, which is then processed through a two-layer regression head to predict 6D pose parameters consisting of a 6D rotation representation {\small$\smash{[r_1 | r_2] \in \mathbb{R}^6}$} and a translation vector {\small$\smash{t \in \mathbb{R}^3}$}. 
For rotation optimization, we use the geodesic distance loss where the rotation output is first mapped to $R \in $ SO(3) via Gram-Schmidt-like orthogonalization~\cite{zhou2019continuity}. The  loss for rotation is as follows:

\begin{equation} \label{eq:3}
l_r(\hat{R}, R_{gt}) = \arccos\left(\frac{\text{tr}(\hat{R} R_{gt}^T) - 1}{2}\right)
\end{equation}
where $\smash{\hat{R}}$ is the predicted rotation matrix obtained by mapping the 6D representation to SO(3), and $\smash{R}_{gt}$ is the ground truth rotation matrix. While it is possible to directly compute the L2 loss between rotation matrices instead of the geodesic loss, both ~\cite{gao20206d} and our experimental results show that the geodesic loss achieves better performance (Details about loss choice can be found in the supplementary). 

For translation, L2 loss is used to measure the distance between the predicted translation $\hat{t}$ and the ground truth translation $t_{gt}$:
\begin{equation} \label{eq:4}
l_t(\hat{t}, t_{gt}) = \| \hat{t}-t_{gt} \|^2_2
\end{equation} 
The overall loss to be optimised is:
\begin{equation} \label{eq:5}
\mathcal{L}_{\phi} = l_r + l_t
\end{equation}

As shown in Figure \ref{fig:four_plots}\subref{fig:a}, while other pre-training objectives such as classification and reconstruction do not much help compared to learning from scratch as in GenPose, the direct regression offers better initial weight parameters of the encoder, accelerating the convergence of end-to-end learning with diffusion.



\paragraph{Joint Training.} 
After the pre-training, we initialize the encoder and regression head with the pre-trained weights and jointly train them alongside the denoising diffusion head. During this stage, the regression head continues to be optimized with the $\smash{L_{\phi}}$, while the diffusion head is trained with the denoising score matching objective defined in Equation \ref{eq:2}:
\begin{small}\begin{equation} \label{eq:6}
\mathcal{L}_{\theta} =  \mathbb{E}_{t}\bigg\{\lambda(t)\mathbb{E}_{\mathbf{x}(0), \\ {\mathbf{x}(t)}}\bigg[ \bigg\| \mathbf{s}_\theta(\mathbf{x}(t),t|O)  + \frac{\mathbf{x}(t)-\mathbf{x}(0)}{\sigma_t^2}\bigg\|_2^2\bigg]\bigg\} 
\end{equation}\end{small}
where now {\small$\smash{\mathbf{x}(0)}$} represents the ground truth 6D pose, {\small$\smash{\mathbf{x}(t)}$} is the noisy pose at time $t$, and {\small$\smash{\mathbf{s}_\theta(\mathbf{x}(t),t|O)}$} is the score function conditioned on the observed point cloud $O$. This joint training allows the encoder to receive simultaneous supervision from the direct pose regression and score-matching loss through backpropagation as 
\begin{equation} \label{eq:5}
\mathcal{L}_{total} = \mathcal{L}_{\phi} + \mathcal{L}_{\theta}.
\end{equation} 



\begin{figure}[h]
  \centering
   \includegraphics[width=0.9\linewidth]{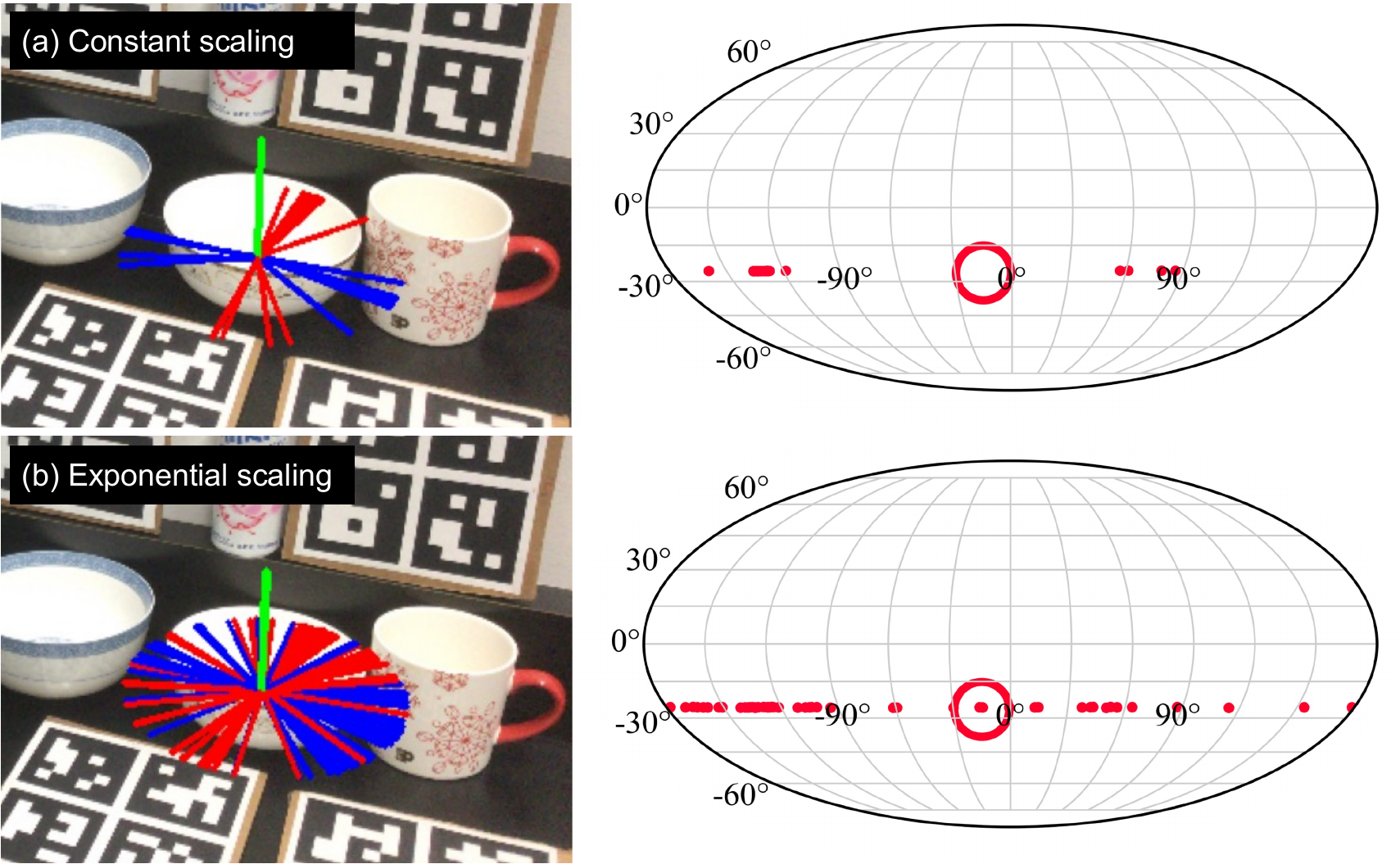}
   \caption{Visualization of sampled poses for a symmetric object under different scaling strategies. Rotation distribution is visualized using Mollweide projection inspired by ~\cite{murphy2021implicit}, where yaw and pitch rotations are mapped to longitude-latitude coordinates, with roll as color. The center of the circle represents the ground truth pose.}
   \label{fig:symmetric_comparison}
  \vspace{-0.3cm}
\end{figure}

\subsection{Score Scaling Guidance}

Given a point cloud condition $O$, the score estimated by our diffusion head represents a vector field that points toward regions of high conditional probability density at each noise level. Specifically, for a noisy pose {\small$\smash{x(t)}$} at time $\smash{t}$, the predicted score {\small$\smash{\mathbf{s}_\theta(\mathbf{x}(t),t|O)}$} provides the direction and magnitude to guide the sample toward higher-density regions of the pose distribution. A straightforward approach to guide a sample toward high-density regions preventing outliers from being sampled would be to uniformly scale the predicted score by a constant factor {\small$\smash{w > 1}$} across all time steps.

However, applying such constant scaling throughout the sampling trajectory strongly pushes samples toward the highest density at all noise levels, leading to mode collapse. In other words, this causes samples to quickly converge to a single mode rather than capturing the full symmetric distribution for objects possessing continuous symmetries (Figure \ref{fig:symmetric_comparison} (top)). 
To preserve the symmetric pose distribution while still improving sample quality, we thus propose a time-dependent score scaling strategy:
\begin{equation} \label{eq:8}
\mathbf{s}_{\theta}^{scaled} = w_t \cdot \mathbf{s}_\theta(\mathbf{x}(t),t|O)
\end{equation} 
The scaling weight {\small$\smash{w_t}$} increases exponentially as $\smash{t}$ approaches 0, following the schedule:
\begin{equation} \label{eq:9}
w_t = w_{min} + (w_{max} - w_{min})\exp(-5t)
\end{equation}
where {\small$\smash{w_{min}}$} and {\small$\smash{w_{max}}$} are hyperparameters which denote the minimum and maximum scaling weights respectively. $\smash{t}$ represents the diffusion time.
It applies minimal guidance at early time steps, allowing thorough exploration of the pose space while gradually increasing the scaling weight as {\small$\smash{t \approx 0}$} to ensure high-quality samples. This simple scheduling  successfully captures the symmetric pose distribution while maintaining pose quality, effectively balancing exploration and exploitation during sampling (Figure \ref{fig:symmetric_comparison} (bottom)). 

\begin{figure}[t]
  \centering
   \includegraphics[width=0.93\linewidth]{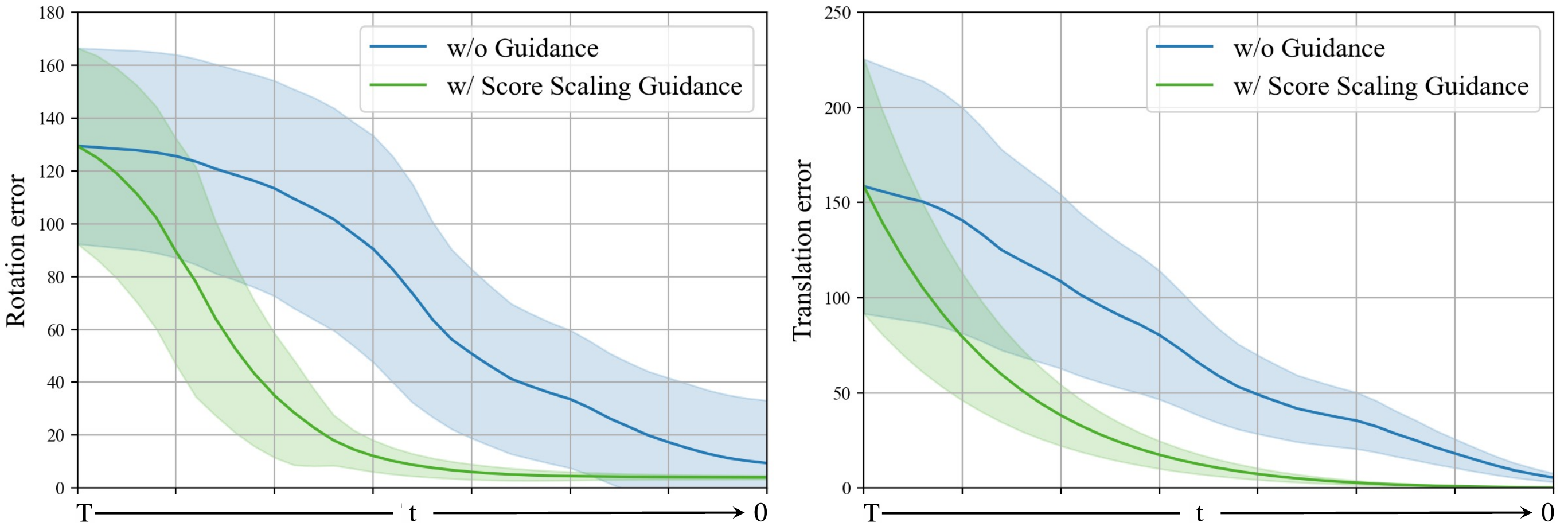}
   \caption{Rotation (left) and translation (right) errors along sampling trajectories from T to 0, comparing our score scaling guidance (green) with baseline (blue). 50 initial noises were sampled.  Solid lines and shaded regions indicate mean and standard deviation, respectively.}
   \label{fig:comare_guidance}
\vspace{-0.4cm}
\end{figure}

Also, Figure \ref{fig:comare_guidance} demonstrates the effectiveness of our score scaling guidance in preventing outliers during sampling. The guided samples (green) show consistently decreasing errors and tighter variance as $\smash{t}$ approaches 0, while samples without guidance (blue) exhibit larger variance and higher mean errors. This empirically validates that our time-dependent score scaling successfully guides samples toward high-density regions.

\section{Experiments}
\label{sec:exp}

\subsection{Implementation}
We use instance masks generated by MaskRCNN~\cite{he2017mask} for a fair comparison with previous methods during inference. Following GenPose~\cite{zhang2023genpose}, our network takes 1024 points as input and is trained with a batch size of 192. We employ the Adam optimizer with a warm-up strategy and exponential learning rate decay scheduling. The same training configuration is applied to both the pre-training and joint training phases. All experiments are conducted on a single NVIDIA RTX 4090 GPU, and our implementation is based on PyTorch~\cite{paszke2019pytorch}.

\begin{figure*}[t]
    \centering
    
    \begin{subfigure}[b]{0.24\textwidth}
        \centering
        \includegraphics[width=\textwidth]{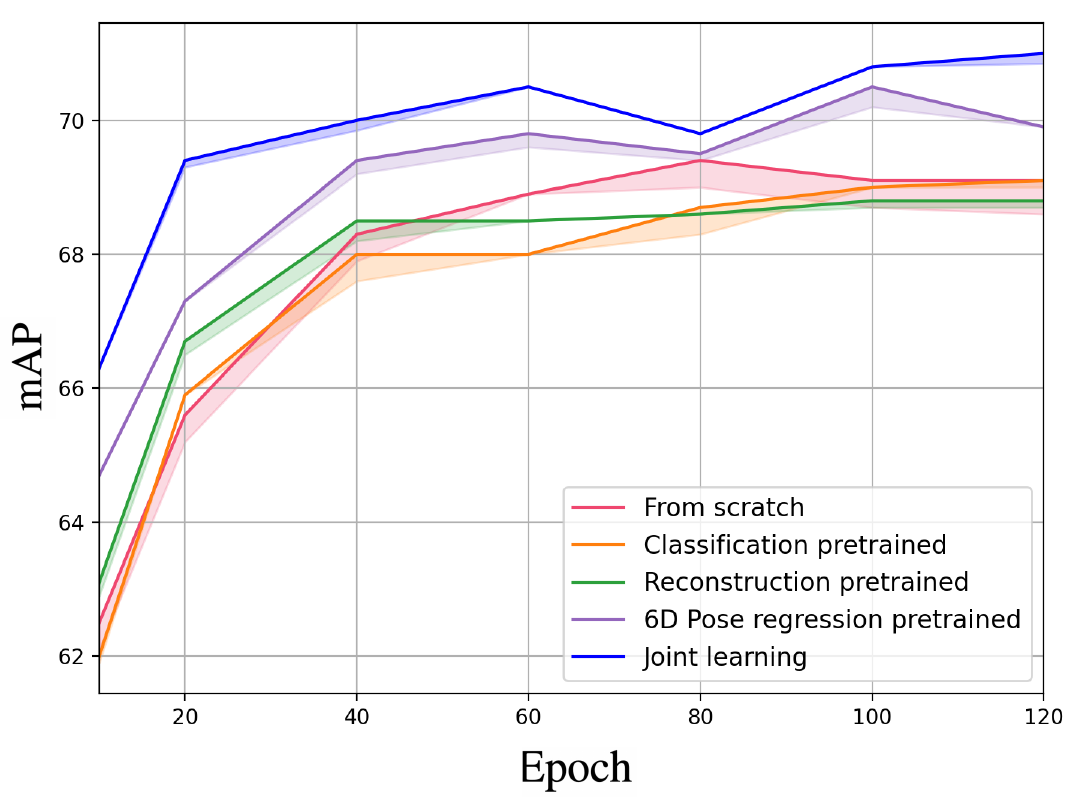}
        \caption{Comparison of pre-training}
        \label{fig:a}
    \end{subfigure}
    \hfill
    \begin{subfigure}[b]{0.24\textwidth}
        \centering
        \includegraphics[width=\textwidth]{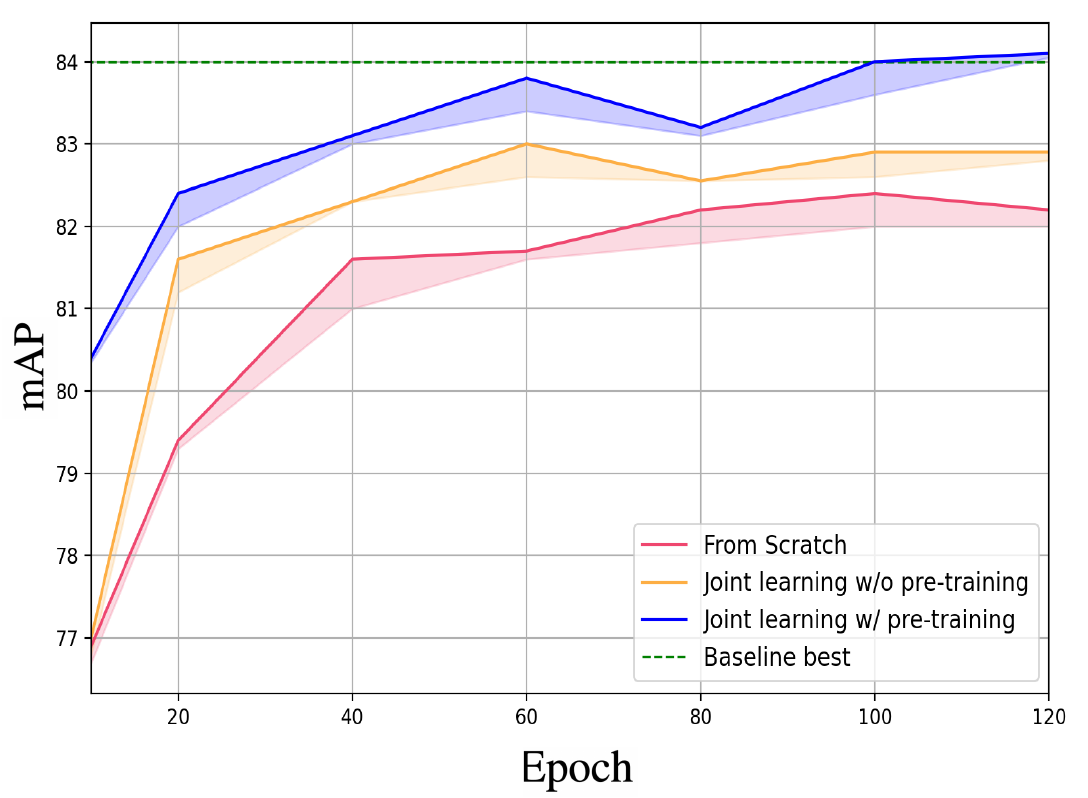}
        \caption{Effectiveness of Joint Learning}
        \label{fig:b}
    \end{subfigure}
    \hfill
    \begin{subfigure}[b]{0.24\textwidth}
        \centering
        \includegraphics[width=\textwidth]{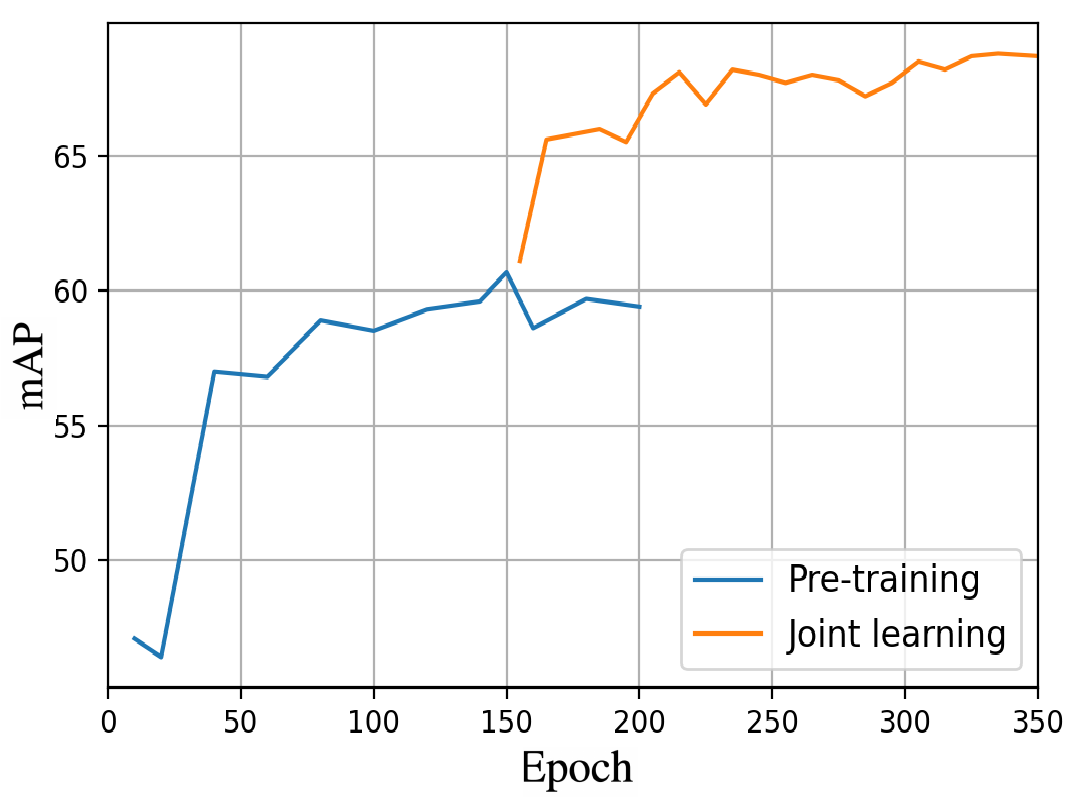}
        \caption{Regression head performance}
        \label{fig:c}
    \end{subfigure}
    \hfill
    \begin{subfigure}[b]{0.24\textwidth}
        \centering
        \includegraphics[width=\textwidth]{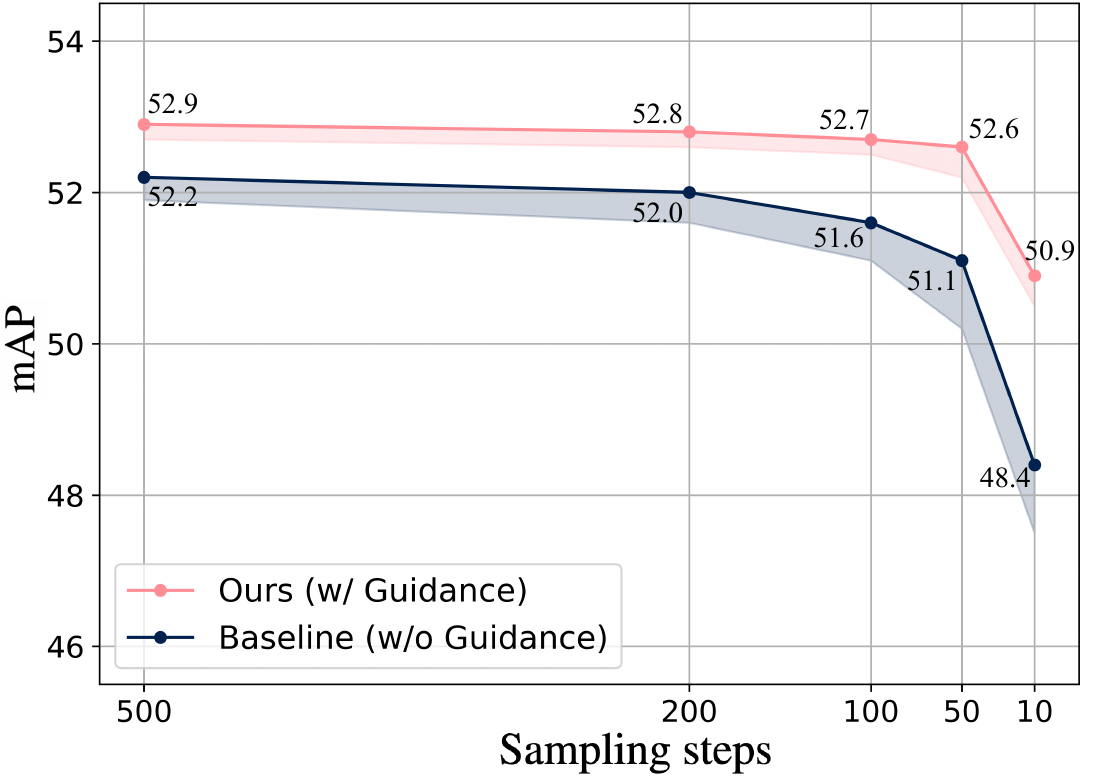}
        \caption{Comparison of sampling steps}
        \label{fig:d}
    \end{subfigure}
    
    \caption{%
    (a) compares various pre-training strategies alongside joint learning, and (b) shows from-scratch vs. joint learning performance with or without 6D pose regression pre-training. (c) illustrates the regression head performance during the pre-training and joint learning phases, and (d) compares performance under different sampling steps. All results are on REAL275, where shaded regions in (a), (b) and (d) indicate min/max over 3 evaluations.}
    \label{fig:four_plots}
     \vspace{-0.4cm}
\end{figure*}


\subsection{Evaluation}
\vspace{-0.2em} 
\paragraph{Datasets.}
To demonstrate the effectiveness of our approach, we conduct comprehensive evaluations on three category-level 6D pose estimation benchmarks: NOCS~\cite{wang2019normalized}, HouseCat6D~\cite{jung2024housecat6d}, and Omni6DPose~\cite{zhang2025omni6dpose}. 

\begin{itemize}
    \item NOCS comprises two datasets: REAL275, a real-world dataset with 4.3K training images and 2.75K testing images across 6 categories (bottle, bowl, camera, can, laptop, and mug). CAMERA25 is a synthetic dataset with 275K training images and 25K testing images in the same categories. Following NOCS, we use both REAL275 and CAMERA25 train datasets to optimize our model. 
    \item HouseCat6D is a real-world dataset featuring photometrically challenging objects, such as glassware and cutlery, with broad viewpoint coverage and complex occlusions. The dataset is divided into 34 training scenes with 20K images, 5 test scenes with 3K images, and 2 validation scenes with 1.4K images. Each training scene includes an average of 6 objects from various categories, while test scenes feature 10 unseen objects per scene.
    \item  Omni6DPose provides the largest benchmark with two training and test datasets: SOPE and ROPE. SOPE comprises 475K synthetic images with 5M annotations across 4,162 instances in 149 categories, while ROPE contains 332K real images with 1.5M annotations across 581 instances in the same categories.

\end{itemize}

\paragraph{Metrics.}
For the REAL275 and HouseCat6D datasets, we report the mean Average Precision (mAP) at three thresholds: $5^{\circ} 2$cm, $10^{\circ} 2$cm, and $10^{\circ} 5$cm where $n^{\circ} m$cm indicates a rotation error within $n$ degrees and a translation error within $m$ centimeters. For ROPE dataset, following ~\cite{zhang2025omni6dpose}, we use the proposed Volume Under Surface (VUS) metric which evaluates pose accuracy across continuous error ranges from $0^{\circ}$ to $n^{\circ}$ for rotation and $0$cm to $m$cm for translation. We report VUS@$5^{\circ} 2$, VUS@$10^{\circ} 2$cm and  VUS@$10^{\circ} 5$cm.




\begin{table}[h]
  \centering
\begin{adjustbox}{scale=0.95}
  \begin{tabular}{l c}
    \toprule
    Method & Total Training time \\
    \midrule
    From Scratch & 11 days \\
    Ours (Full convergence) & 7.5 days \\
    Ours (To baseline best at $10^{\circ}5cm$) & \textbf{0.5 days} \\
    \bottomrule
  \end{tabular}
  \end{adjustbox}
  \caption{Comparison of Training Time on NOCS dataset between baseline (From Scratch) and Joint Learning.}
  \label{tab:training_time}
  \vspace{-0.2em}
\end{table}

\vspace{-0.7em}
\subsection{Comparison with Baseline}
\begin{figure}[hb!]
 \centering
 \includegraphics[width=0.95\linewidth]{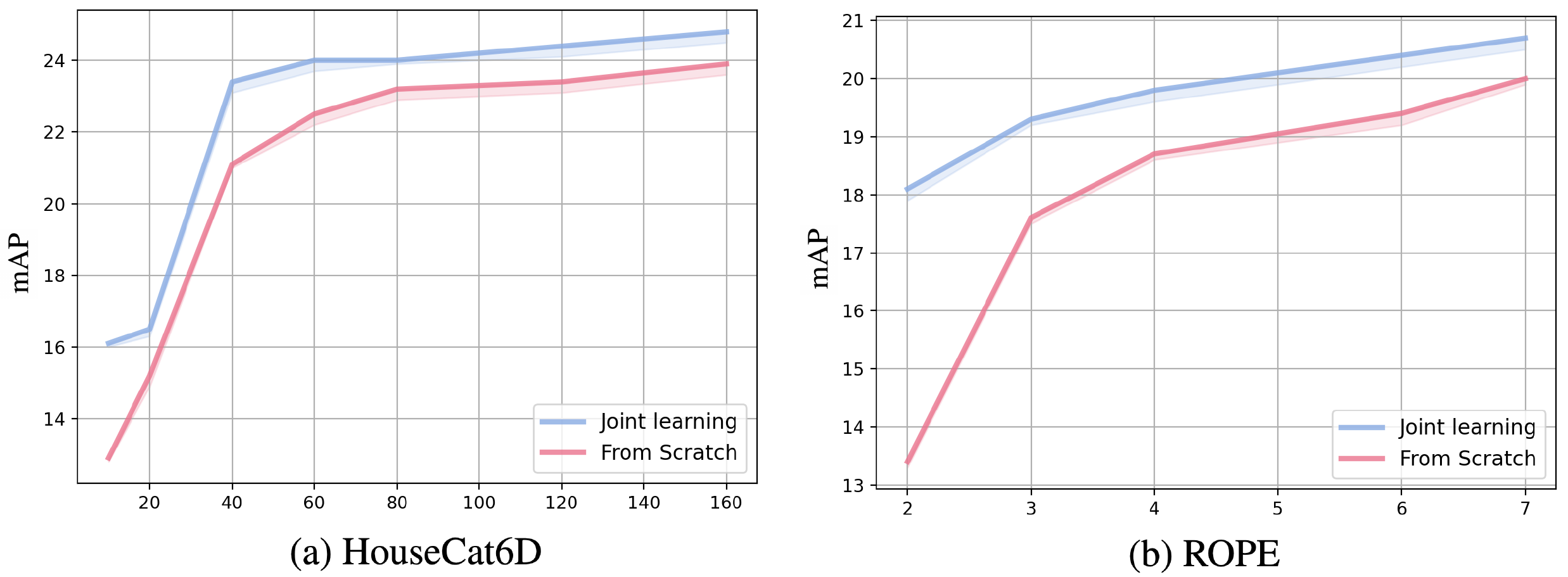}
 \caption{Comparison between training from scratch and joint learning  on HouseCat6D and ROPE datasets on $10^\circ 5\text{cm}$.}
 \label{fig:housecat_rope}
\vspace{-0.5em}
\end{figure}

\noindent \textbf{Efficiency of Joint learning in model convergence.}
We first evaluate the efficiency of our joint learning strategy against the baseline approach of training from scratch. All experiments are conducted on a single NVIDIA RTX 4090 GPU with identical hyperparameter settings. As shown in Table \ref{tab:training_time}, the model converges in 7.5 days with our Joint learning method, representing a 1.5$\times$ speedup compared to the baseline. Figure \ref{fig:four_plots}\subref{fig:b} further demonstrates this efficiency gain, showing that our approach (blue line) reaches and surpasses the baseline's best performance (dashed green line) within approximately 100 epochs (about 0.5 days) in $10^\circ 5\text{cm}$ mAP threshold, while the baseline method (red line) requires significantly more training time to achieve comparable results. Figure \ref{fig:housecat_rope} extends these findings to HouseCat6D and ROPE datasets, where joint learning consistently achieves faster convergence and better performance.

\noindent \textbf{Inference efficiency.} Table {\ref{tab:baseline_compare}} compares the inference time (FPS) between our method and GenPose. $\smash{K}$ indicates the number of sampled pose candidates, and ``Ranker" denotes whether the additional EnergyNet~\cite{zhang2023genpose} is used for outlier filtering. When ``Ranker" is not used, poses are randomly selected with {\small$\smash{\delta=60\%}$} for mean pooling. If {\small$\smash{K=1}$}, the single sampled pose is taken as the final prediction. In single-frame estimation, reducing pose candidates K from 50 to 1 improves speed from 6.20 to 8.21 FPS, with moderate gains due to batch-wise parallel sampling on GPU. In a more practical pose tracking scenario, following GenPose~\cite{zhang2023genpose} where well-initialized values from previous frames are leveraged, we achieve a more substantial speedup (20.6→36.1 FPS), as our single-pose inference eliminates both the need for 50 candidates sampling and an additional ranking network for pose verification. By removing this ranking network, the needed number of paramaters is halved over GenPose (4.4→2.2M).

\begin{table}
\begin{adjustbox}{scale=0.73}
  \centering
  \small
  \begin{tabular}{l |c c | c c c c}
    \toprule
    Method  & K & Ranker & $5^\circ 2\text{cm}\uparrow$  & $10^\circ 2\text{cm}\uparrow$ & Speed(FPS)$\uparrow$ & Params(M)$\downarrow$\\
    \midrule
    GenPose$^1$  & 1 & \tikzxmark & {13.5 / 29.7} & {21.5 / 40.4} & \textbf{8.25 / 36.8}  & \textbf{2.2}\\
    GenPose$^2$  & 20 & \tikzxmark & {47.9 / 59.0} & {66.6 / 70.6} & {7.67 / 31.8}  & \textbf{2.2}\\
    GenPose$^3$  & 50 & \tikzxmark & {49.4 / 60.9} & {68.5 / 78.2} & {6.97 / 25.7}  & \textbf{2.2}\\
    GenPose$^4$  & 20 & \tikzcmark & {50.8 / 62.8} & {71.3 / 80.3} & {7.04 / 27.2}  & {4.4}\\
    GenPose$^5$  & 50 & \tikzcmark & {52.1 / 63.5} & {72.4 / 81.4} & {6.20 / 20.6}  & {4.4}\\
    \midrule
    Ours (J)  & 50 & \tikzxmark & {50.8 / 61.6}  & {69.7 / 79.5} & {7.03 / 25.6} & \textbf{2.2} \\
    Ours (G)  & 1 & \tikzxmark & {52.2 / 63.1} & {72.2 / 81.6} & {8.21 / 36.1}  & \textbf{2.2}\\
    Ours (J+G)  & 1 & \tikzxmark & \textbf{52.6} / \textbf{64.6} & \textbf{73.4} / \textbf{82.2} & {8.23 / 36.4}  & \textbf{2.2}\\
    \bottomrule
\end{tabular}
\end{adjustbox}
\caption{Comparison of baseline and our methods (J: joint learning, G: score scaling guidance) with different numbers of sampled pose candidates K on REAL275 dataset. The left and right sides of ‘/’ respectively indicate \textbf{single frame pose estimation} and \textbf{pose tracking results.}}
\label{tab:baseline_compare}
\vspace{-0.5cm}
\end{table}

\noindent \textbf{Accuracy.} Table \ref{tab:baseline_compare} also presents a comprehensive evaluation of our proposed methods with GenPose on the REAL275 dataset, demonstrating substantial improvements in performance. 
Following the baseline's setteing but excluding ``Ranker", our joint learning approach (J) consistently outperforms the diffusion-only training strategy of GenPose$^3$, highlighting the effectiveness of simultaneously training both regression and diffusion heads. 
Furthermore, by incorporating score scaling guidance, Ours (G) enables reliable single pose estimation while significantly reducing computational cost - achieving higher accuracy than GenPose$^1$ without requiring its 50 pose samples and additional filtering network. This is particularly notable given that the single pose sampling attempt in GenPose$^1$ suffers from frequent outliers. Our combined approach (G+J) achieved the best performance.



\begin{table*}[h]
\centering
\small
\caption{Quantitative comparison on REAL275 and CAMERA25. $^\dagger$Results based on probabilistic method.}
\begin{subtable}{0.49\textwidth}
\raggedright{
\caption{REAL275 dataset.}
\label{tab:real275}
  \begin{adjustbox}{scale=0.62}
  \begin{tabular}{l | c c c | c c c}
    \toprule
    METHOD & Input & Prior & K & $5^\circ 2\text{cm}\uparrow$ & $10^\circ 2\text{cm}\uparrow$ & Params(M)$\downarrow$\\
    \specialrule{.12em}{.05em}{.05em} 
    NOCS~\cite{wang2019normalized} & RGB-D & \tikzxmark & 1 & 7.2 & 13.8 & {-} \\
    DualPoseNet~\cite{lin2021dualposenet} & RGB-D & \tikzxmark & 1 & 29.3 & 50.0 & {67.9} \\
    VI-Net~\cite{lin2023vi} & RGB-D & \tikzxmark & 1 & 50.0 & 70.8  & {27.3} \\
    AG-Pose~\cite{lin2024instance} & RGB-D & \tikzxmark & 1 & 54.7 & \textbf{74.7}& {207.6} \\
    SPD~\cite{tian2020shape} & RGB-D & \tikzcmark & 1 & 19.3 & 43.2 & {18.3} \\
    CR-Net~\cite{wang2021category} & RGB-D & \tikzcmark & 1 & 27.8 & 47.2 & {-} \\
    SecondPose~\cite{chen2024secondpose} & RGB-D & \tikzcmark & 1 & \textbf{56.2} & \textbf{74.7} & {33.6} \\
    \midrule
    GPV-Pose~\cite{di2022gpv} & D & \tikzxmark & 1 & 32.0 & 50.0 & {-} \\
    SSP-Pose~\cite{zhang2022ssp} & D & \tikzcmark & 1 & 34.7 & - & {-} \\
    RBP-Pose~\cite{zhang2022rbp} & D & \tikzcmark & 1 & 38.2 & 63.1 & {-} \\
    HS-Pose~\cite{zheng2023hs} & D & \tikzxmark & 1 & 46.5 & 68.6  & {6.1} \\
    GenPose$^\dagger$~\cite{zhang2023genpose} & D & \tikzxmark & 50 & 52.1 & 72.4 & \textbf{2.2} \\
    {GenPose$^\dagger$}~\cite{zhang2023genpose} & D & \tikzxmark & 1 & 13.5 & 21.5 & \textbf{2.2} \\
    \midrule
    {Ours$^\dagger$ (G)} & D & \tikzxmark &  1 & {52.2} & {72.2} & \textbf{2.2}\\
    {Ours$^\dagger$ (J)} & D & \tikzxmark &  50 & {50.8} & {69.7} & {4.4}\\
    {Ours$^\dagger$ (G+J)} & D & \tikzxmark &  1 & \textbf{52.6} & \textbf{73.4} & \textbf{2.2} \\
    \midrule
\end{tabular}
\end{adjustbox}}
\end{subtable}
\begin{subtable}{0.49\textwidth}
\raggedleft{
\caption{CAMERA25 dataset.}
\label{tab:camera}
\small
\begin{adjustbox}{scale=0.66}
\begin{tabular}{l | c c c | c c c}
    \toprule
    METHOD & Input & Prior & K & $5^\circ 2\text{cm}\uparrow$ & $10^\circ 2\text{cm}\uparrow$ & Params(M)$\downarrow$\\
    \specialrule{.12em}{.05em}{.05em} 
    NOCS~\cite{wang2019normalized} & RGB-D & \tikzxmark & 1 & 32.3 & 48.2 & {-} \\
    DualPoseNet~\cite{lin2021dualposenet} & RGB-D & \tikzxmark & 1 & 64.7 & 77.2 & {67.9}\\
    VI-Net~\cite{lin2023vi} & RGB-D & \tikzxmark & 1 & 74.1 & 79.3 & {27.3}  \\
    AG-Pose~\cite{lin2024instance} & RGB-D & \tikzxmark & 1 & \textbf{77.8} & \textbf{85.5} & {207.6} \\
    SPD~\cite{tian2020shape} & RGB-D & \tikzcmark & 1 & 54.3 & 73.3 & {18.3} \\
    CR-Net~\cite{wang2021category} & RGB-D & \tikzcmark & 1 & 72.0 & 81.0 & {-}\\
    \midrule
    GPV-Pose~\cite{di2022gpv} & D & \tikzxmark & 1 & 72.1 & -  & {-} \\
    SSP-Pose~\cite{zhang2022ssp} & D & \tikzcmark & 1 & 64.7 & - & {-} \\
    RBP-Pose~\cite{zhang2022rbp} & D & \tikzcmark & 1 & 73.5 & 82.1 & {-} \\
    HS-Pose~\cite{zheng2023hs} & D & \tikzxmark & 1 & 73.3 & 80.4  & {6.1} \\
    GenPose$^\dagger$~\cite{zhang2023genpose} & D & \tikzxmark & 50 & 79.9 & 84.6 & \textbf{2.2}\\
    {GenPose$^\dagger$}~\cite{zhang2023genpose} & D & \tikzxmark & 1 & {32.8} & {35.9} & \textbf{2.2} \\

    \midrule
    {Ours$^\dagger$ (G)} & D & \tikzxmark &  1 & {80.0} & {84.8} & \textbf{2.2} \\
    {Ours$^\dagger$ (J)} & D & \tikzxmark &  50 & {78.7} & {84.2}& {4.4} \\
    {Ours$^\dagger$ (G+J)} & D & \tikzxmark &  1 & \textbf{80.4} & \textbf{85.1}& \textbf{2.2} \\
    \midrule
  \end{tabular}
    \end{adjustbox}}
    \end{subtable}
\vspace{-1em}
\end{table*}

\begin{figure}[h]
  \centering
   \includegraphics[width=0.9\linewidth]{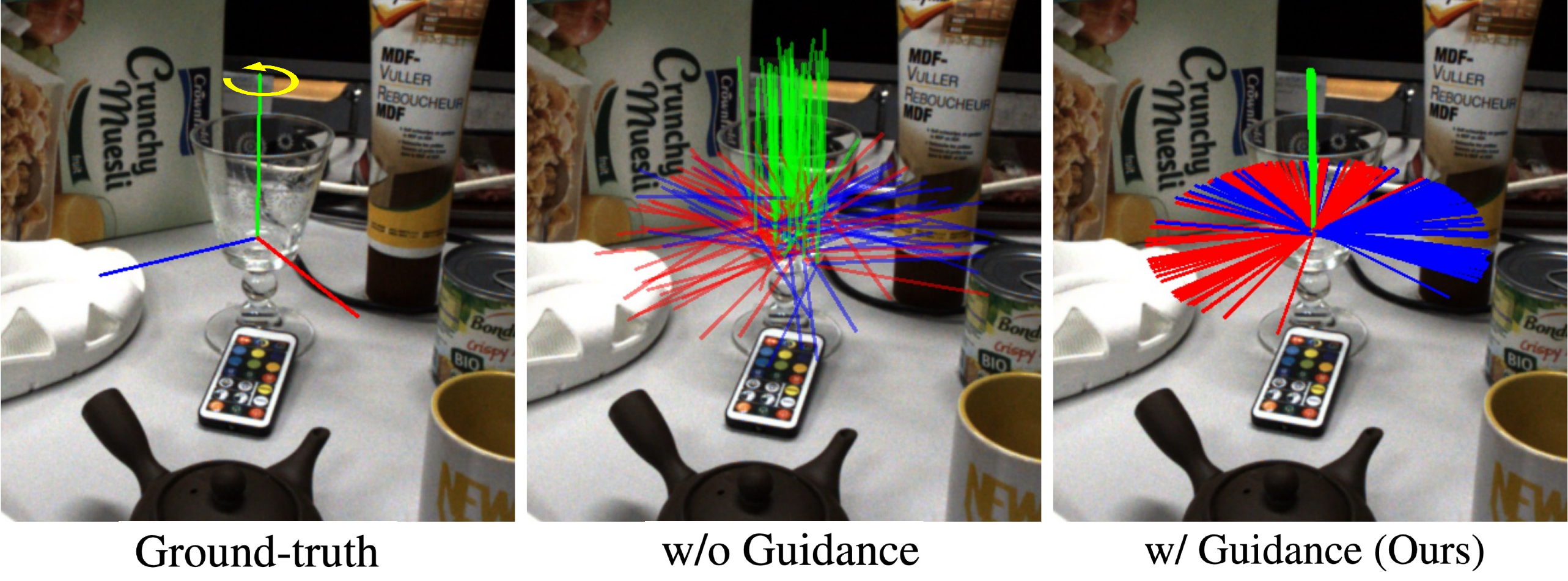}
   \caption{Qualitative results of score scaling guidance on `Glass' category object from HouseCat6D dataset.}
   \label{fig:glass_result}
   \vspace{-1em}
\end{figure}

\subsection{Comparison with State-of-the-Art}

Table \ref{tab:real275} and Table \ref{tab:camera} show quantitative comparisons with other state-of-the-art methods on the REAL275 dataset and CAMERA25 dataset respectively. Among depth-only methods, our method with both joint learning and  score scaling guidance achieves superior performance across all metrics even with single sample inference, while maintaining the lowest parameters compared to others. 

While SecondPose~\cite{chen2024secondpose} achieves competitive results, it relies on both RGB and depth modalities as well as semantic priors from DINOv2. In contrast, our method demonstrates strong performance using only depth input, without additional priors. As a generative approach, it further benefits from the potential ability to capture multi-modal distributions for ambiguous objects, such as symmetric or partially occluded instances, where multiple valid poses may exist. This ability makes our method more robust in real-world scenarios with common pose ambiguities while still achieving competitive performance compared to other methods. More results are found in the supplementary.


\subsection{Robustness of Score Scaling Guidance}
We further validate the robustness and generalizability of our score scaling guidance through experiments on challenging real-world scenarios. 

As shown in Table \ref{tab:housecat_and_rope}, our depth-only method achieves a performance comparable to the RGB-D-based methods on the HouseCat6D dataset, which contains numerous challenging cases such as transparent objects shown in Figure \ref{fig:glass_result}. Moreover, Table \ref{tab:housecat_and_rope} shows that our guidance strategy effectively generalizes to multi-modal frameworks - when integrated into GenPose++~\cite{zhang2025omni6dpose} which utilizes both RGB and depth inputs on the ROPE dataset. These results demonstrate that our Score Scaling Guidance offers a robust solution with significant potential to enhance diffusion-based pose estimation across various real-world scenarios and input modalities.

Figure \ref{fig:four_plots}\subref{fig:d} illustrates the robustness of our method, where we replaced the adaptive-step PF-ODE solver with a DDIM sampler~\cite{song2020denoising} to enable explicit sampling step control. Our method maintains consistent performance even with drastically reduced sampling steps, while the baseline performance decreases significantly at 10 steps. This validates that our score scaling method effectively accelerates convergence to the target conditional distribution. Additional experiments with the DDIM sampler can be found in the supplementary material.

\begin{figure}[h]
  \centering
   \includegraphics[width=1.0\linewidth]{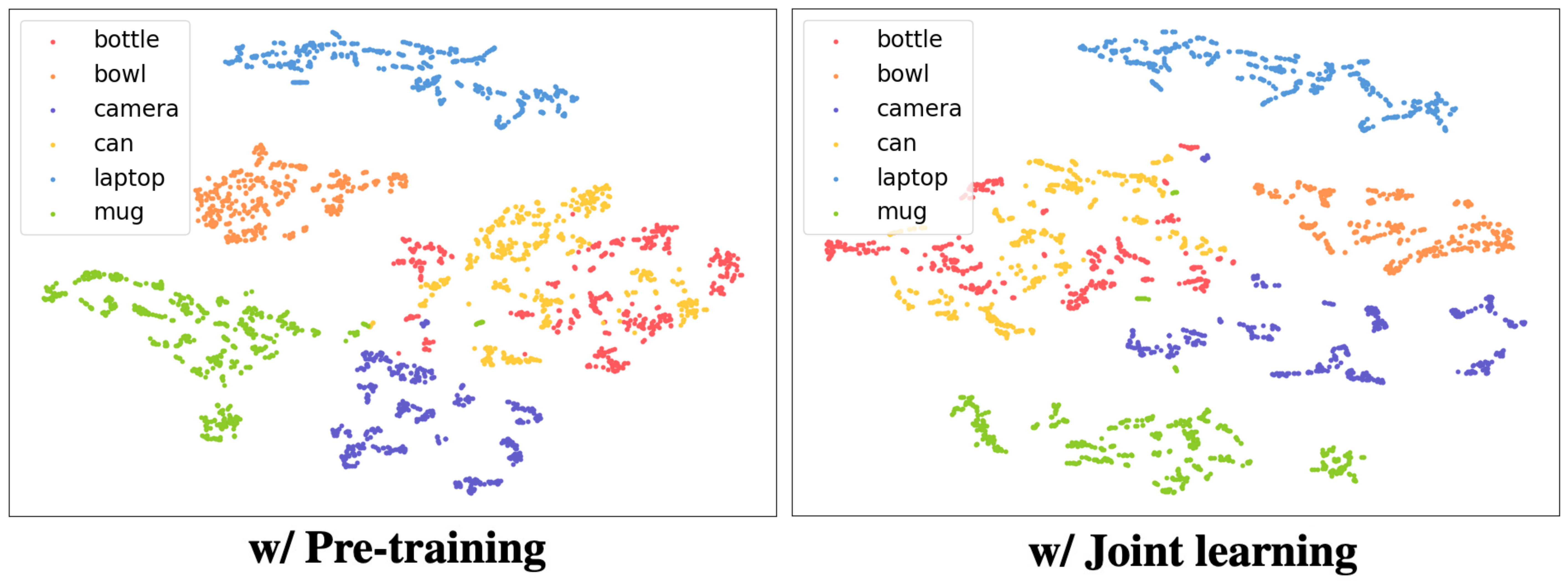}
   \caption{Feature visualization.}
   \label{fig:feature}
    \vspace{-0.4cm}
\end{figure}

\subsection{Ablation Studies}
\paragraph{Effectiveness of Pre-training Regression head.} Figure \ref{fig:four_plots}\subref{fig:b} demonstrates the advantage of pre-training the encoder and the regression head before joint learning. The pre-trained model (blue line) performs better than training from scratch (yellow line) throughout the training. This improvement stems from the reduced optimization complexity of the encoder, as pre-training establishes effective 6d pose estimation features before addressing the joint learning objective. This is further supported by Figure \ref{fig:feature}, which reveals consistent feature distribution patterns between pre-training and joint learning phases. Without pre-training, the encoder optimize for both regression and diffusion loss from scratch, making the learning process more challenging. Moreover, Figure \ref{fig:four_plots}\subref{fig:c} shows that joint learning not only benefits the diffusion head but also boosts the regression head's performance. Upon transitioning to joint learning(orange line), the regression head shows a significant performance improvement compared to pre-training alone. This suggests a mutually beneficial interaction between the regression and diffusion objectives during joint optimization.


\begin{figure}[h]
\vspace{-0.3em}
  \centering
   \includegraphics[width=0.9\linewidth, height=0.3\linewidth]{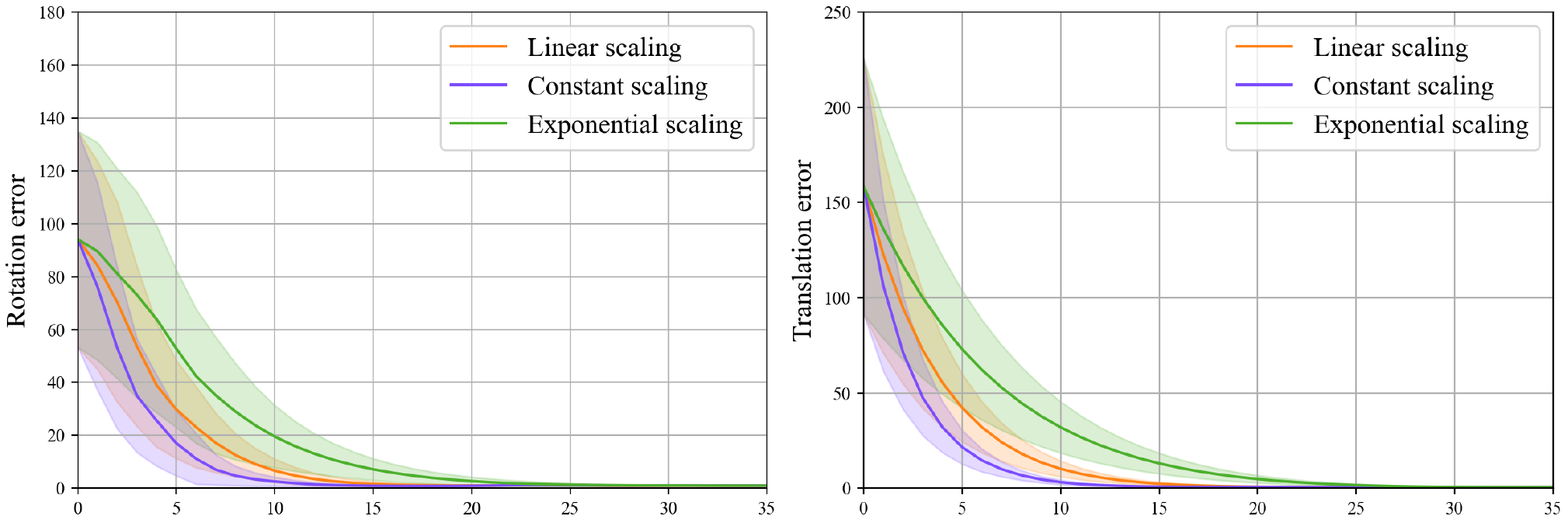}
   \caption{Comparison of rotation (left) and translation (right) errors for symmetric objects under different scaling strategies (linear, constant, and exponential).}
   \label{fig:weight_compare}
   \vspace{-0.7cm}
\end{figure}

\paragraph{Guidance in Symmetric Objects}

\begin{figure}[h]
  \centering
   \includegraphics[width=0.95\linewidth]{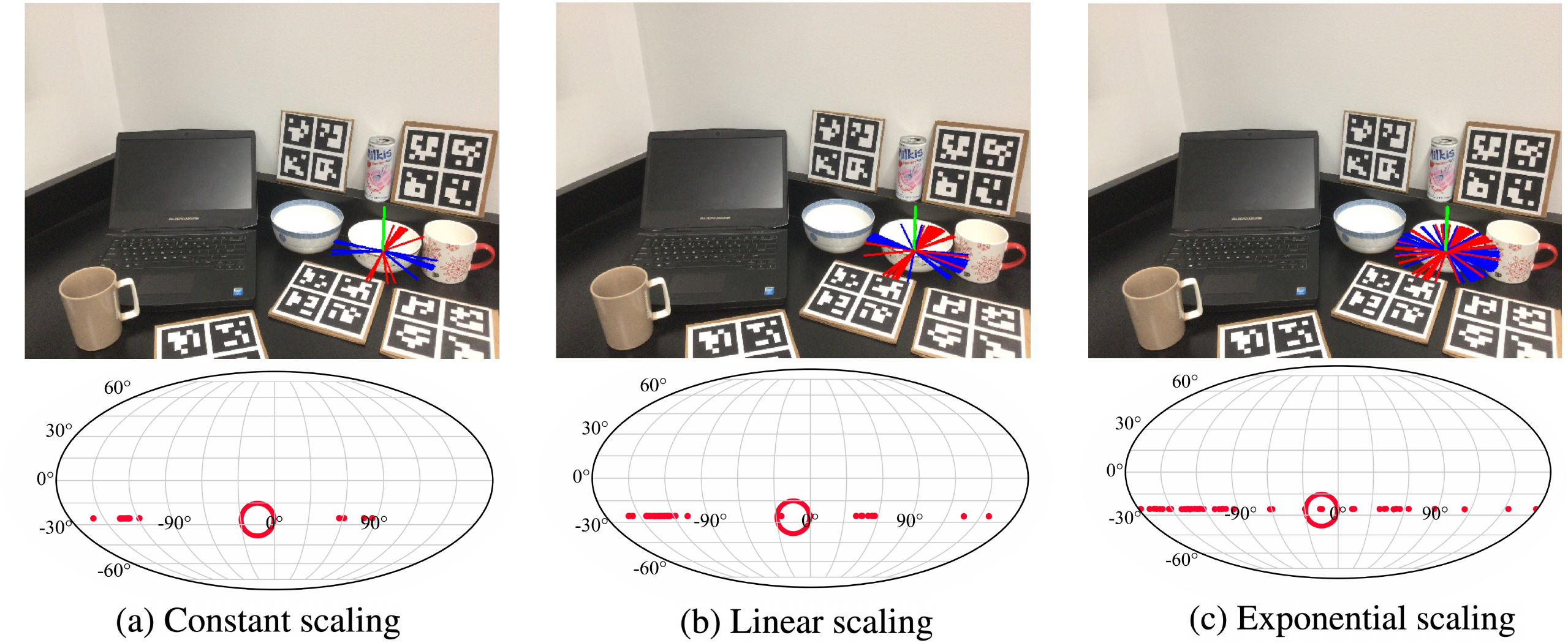}
   \caption{Effect of score scaling weight schedules.}
   \label{fig:sym_schedule}
   \vspace{-0.3cm}
\end{figure}

We analyze sampling trajectories and final pose distributions across different scaling schedules to examine how our time-dependent score scaling affects symmetric objects. As shown in Figure \ref{fig:weight_compare}, applying stronger guidance in early steps (constant \textgreater linear \textgreater exponential) accelerates convergence toward ground truth poses, with all approaches ultimately achieving comparable accuracy (approximately 1$^\circ$ rotation and 0.5cm translation error). However, Figure \ref{fig:sym_schedule} unveils a crucial trade-off in this process: although constant scaling yields low error metrics, it prematurely forces samples to converge to a single symmetric configuration, resulting in mode collapse and failing to capture the inherent multi-modal distribution of symmetric objects. In contrast, exponential scaling not only maintains the diverse range of valid pose modes but also achieves similar accuracy, thereby offering an optimal balance between precise pose estimation and faithful representation of the underlying distribution.

\begin{table}[h]
\centering
\begin{adjustbox}{scale=0.65}
  \small
    \begin{tabular}{l | c c c | c c c}

    \specialrule{.12em}{.05em}{.05em} 
    {\textbf{HouseCat6D}} & Input & Prior & K &  $5^\circ 2\text{cm} \uparrow$ & $10^\circ 2\text{cm} \uparrow$ & Params(M)$\downarrow$  \\
    \specialrule{.12em}{.05em}{.05em} 
    VI-Net~\cite{lin2023vi} & RGB-D & \tikzxmark & 1 & 8.4 & 20.5 & {27.3} \\
    SecondPose~\cite{chen2024secondpose} & RGB-D & \tikzcmark & 1 & \textbf{11.0} & \textbf{25.3} & {33.6} \\
    \midrule
    FS-Net~\cite{chen2021fs} & D & \tikzxmark & 1 & 3.3 & 17.1 & {41.2}  \\
    GPV-Pose~\cite{di2022gpv} & D & \tikzxmark & 1 & 3.5 & 17.8 & {-}  \\
    GenPose$^\dagger$~\cite{zhang2023genpose} & D & \tikzxmark & 50 & 8.5 & 25.3  & {4.4} \\
    GenPose$^\dagger$~\cite{zhang2023genpose} & D & \tikzxmark & 1 & {0.3} & {1.7}  & \textbf{2.2} \\
    \midrule 
    {Ours$^\dagger$ (G)} & D & \tikzxmark & 1 & \textbf{8.9} & \textbf{26.2} & \textbf{2.2} \\
    \midrule    
    {} & {} & {} & {} & {} & {}  \\
    \specialrule{.12em}{.05em}{.05em} 
    {\textbf{ROPE}} & Input & Prior & K & VUS@$5^\circ 2\text{cm} (\uparrow)$ &VUS@$10^\circ 2\text{cm} (\uparrow)$& Params(M)$\downarrow$  \\
    \specialrule{.12em}{.05em}{.05em} 
    SGPA~\cite{chen2021sgpa} & RGB-D & \tikzcmark & 1 & 4.3 & 9.3  & {-} \\
    NOCS~\cite{wang2019normalized} & RGB-D & \tikzxmark & 1 & 0.0 & 0.0 & {-}  \\
    IST-Net~\cite{liu2023net} & RGB-D & \tikzxmark & 1 & 2.0  & 5.3 & {21}  \\
    GenPose++$^\dagger$~\cite{zhang2025omni6dpose} & RGB-D & \tikzxmark & 50 & 10.0 & 19.5  & {4.4} \\
    GenPose++$^\dagger$~\cite{zhang2025omni6dpose} & RGB-D & \tikzxmark & 1 & {1.9} & {3.9}  & \textbf{2.2} \\
    \midrule
    { Ours$^\dagger$ (G)} & RGB-D & \tikzxmark & 1 & \textbf{10.9} & \textbf{20.8} & \textbf{2.2}  \\
    \midrule    

  \end{tabular}
  \end{adjustbox}
  \caption{Quantitative comparison between our Score Scaling Guidance and other baselines on HouseCat6D and ROPE dataset. $^\dagger$Results based on probabilistic method.}
  \label{tab:housecat_and_rope}
  \vspace{-0.55cm}
\end{table}
\section{Conclusion}
\label{sec:conc}
\vspace{-0.55em}
In this paper, we tackle key challenges in framing category-level 6D pose estimation as a generative modeling task. Our approach enhances both training efficiency and pose accuracy by introducing a joint learning strategy that combines direct pose regression with diffusion score matching objectives. Moreover, we propose exponential score scaling guidance that effectively eliminates outlier samples while preserving multi-modal characteristics of symmetric objects, removing the need for additional evaluation networks. Through extensive experiments on existing benchmarks, we demonstrate that our simple yet powerful method achieves state-of-the-art performance even with single-sample inference, making it both efficient and practical for real-world applications.

\section*{Acknowledgements}
This work was supported by NST grant (CRC 21011, MSIT), IITP grant (RS-2023-00228996, RS-2024-00459749, RS-2025-25443318, RS-2025-25441313, MSIT) and KOCCA grant (RS-2024-00442308, MCST).
{
    \small
    \bibliographystyle{ieeenat_fullname}
    \bibliography{main}
}

\clearpage
\clearpage
\setcounter{page}{1}
\newcounter{counter}[section]
\setcounter{section}{0}

\setcounter{figure}{0}
\setcounter{table}{0}
\setcounter{equation}{0}

\maketitlesupplementary

\makeatletter
\renewcommand{\thesection}{S.\arabic{section}}
\renewcommand{\thefigure}{S\arabic{figure}}
\renewcommand{\thetable}{S\arabic{table}}
\renewcommand{\thepage}{\arabic{page}}

In this supplementary document, we first provide implementation details of our method (Section {\ref{supp:imp}}), followed by extensive experimental analyses that complement the main paper (Section \ref{sec:analysis}).  Lastly, we show additional qualitative results that demonstrate our method's effectiveness across various object categories and challenging scenarios. (Section \ref{sec:qual}).



\section{Implementation Details}
\label{supp:imp}

\paragraph{Architecture.} Based on the architecture of GenPose~\cite{genpose}, our network consists of PointNet++~\cite{pointnetpp}, which extracts a 1024-dimensional global feature from the input point cloud. The time step \smash{$t$} and noisy pose \smash{$p(t)$} are embedded through MLPs to produce 128-dimensional and 256-dimensional feature vectors, respectively. These three features are concatenated to a 1408-dimensional vector and fed into the denoising diffusion head to predict a 9D score vector (6D rotation representation and 3D translation). The regression head consists of MLP of size 1024$\times$512$\times$512$\times$9, directly predicting the 6D representation and 3D translation vectors.

\paragraph{6D Rotation Representation.} For rotation representation, we employ the continuous 6D representation following ~\cite{genpose, continuity}, where $g_{GS}$ maps SO(3) to 6D representation by retaining the first two columns of the rotation matrix:
\begin{equation} \label{eq:supp1}
 g_{GS}\left(\begin{bmatrix} a_1 & a_2 & a_3 \end{bmatrix}\right) = \begin{bmatrix} a_1 & a_2 \end{bmatrix}
\end{equation} 
$f_{GS}$ maps 6D representation back to SO(3) through Gram-Schmidt-like orthogonalization~\cite{continuity}:

\begin{equation} \label{eq:supp2}
f_{GS}\left(\begin{bmatrix}  a_1 & a_2 \\\end{bmatrix}\right) = \begin{bmatrix}  b_1 & b_2 & b_3 \\ \end{bmatrix}
\end{equation} 
\begin{equation} \label{eq:supp3}
b_i = \begin{bmatrix}\begin{cases} N(a_1) & \text{if } i = 1 \\ N\big(a_2 - (b_1 \cdot a_2)b_1\big) & \text{if } i = 2 \\ b_1 \times b_2 & \text{if } i = 3 \end{cases}\end{bmatrix}^\top
\end{equation} 

\begin{figure}[h]
  \centering
   \includegraphics[width=0.85\linewidth]{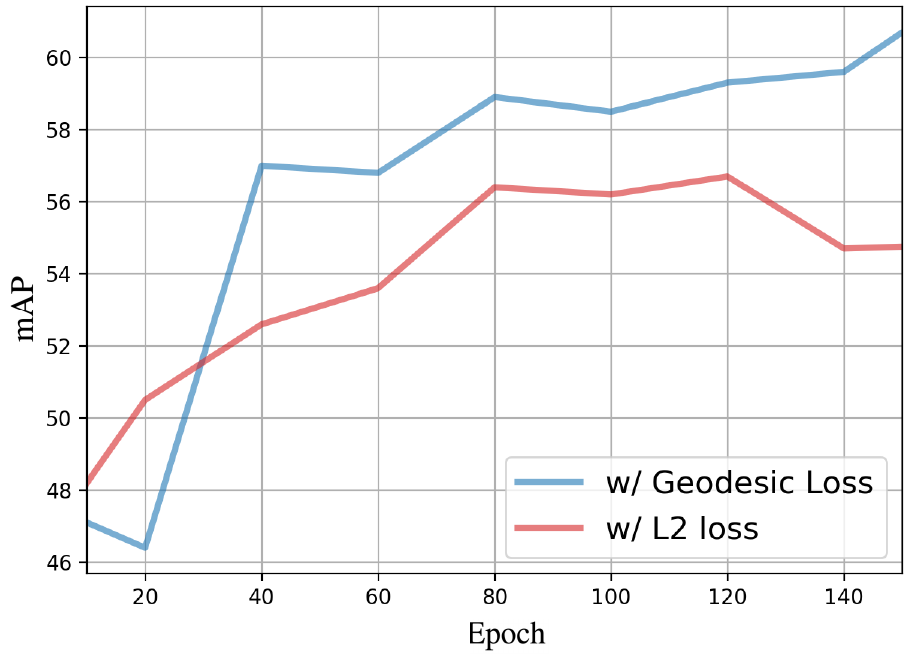}
   \caption{Comparison of pre-training performance on REAL275 dataset using different rotation loss functions for the regression head.}
   \label{fig:rotation_loss}
\end{figure}

\paragraph{Training Details.}
In the pre-training phase, we train only the PointNet++ encoder and regression head, where the regression head outputs 9D vectors (6D rotation and 3D translation). The network is optimized by first mapping the predicted 6D rotation to SO(3) to compute the geodesic loss and combining it with L2 loss for translation. We adopt the geodesic distance for rotation loss, as it provides a clearer learning goal on SO(3) compared to L2 loss~\cite{geodesic}. Also, as shown in Figure \ref{fig:rotation_loss}, our experiments demonstrate superior performance with geodesic loss (blue) compared to L2 loss (red). For objects that are fully symmetric around the y-axis (bottle, bowl, and can) in the NOCS dataset~\cite{nocs}, we compute the rotation loss only with the y-axis to account for their symmetry properties. In the joint learning phase, we initialize the encoder and regression head with the pre-trained weights and simultaneously train both regression and diffusion heads. The regression head maintains its pre-training loss function, while the diffusion head is optimized with the score-matching objective. The network is then updated using the sum of these two losses.

\section{More Results and Analyses}
\label{sec:analysis}

\subsection{Inference Efficiency with DDIM Sampler}

Diffusion-based methods inherently suffer from computational burden due to multiple sampling steps required during inference. To address this limitation and enable explicit control over the number of sampling steps, we replace the original adaptive-step PF-ODE solver with a deterministic DDIM sampler~\cite{ddim}, as mentioned in the main paper (Figure \ref{fig:four_plots}\subref{fig:d}). We perform sampling using the following DDIM update equation:


\begin{equation} \label{eq:ddim}
\begin{aligned}
\mathbf{x}_{t-1} &= \mathbf{x}_t + \sigma_t^2 \, (w_t \cdot \mathbf{s}_\theta(\mathbf{x}_t, t)) \\
&\quad + \sqrt{\sigma_{t-1}^2 - (\eta \sigma_{t-1})^2} \cdot (-\mathbf{s}_\theta(\mathbf{x}_t, t) \cdot \sigma_t) \\
&\quad + \eta \sigma_{t-1} \boldsymbol{\epsilon}_t
\end{aligned}
\end{equation}

where $\mathbf{x}_t$ and $\mathbf{x}_{t-1}$ represent a generated pose samples at time steps {\small\smash{$t$}} and {\small\smash{$t-1$}} respectively, $w_t$ is our proposed score scaling weight, $\sigma_t$ represents the noise level at time step $t$, {\small\smash{$\mathbf{s}_\theta(\mathbf{x}_t, t)$}} is the learned score function, $\eta$ controls the stochasticity of sampling (set to 0 for deterministic sampling), and {\small\smash{$\boldsymbol{\epsilon}_t \sim \mathcal{N}(0, \mathbf{I})$}} is Gaussian noise.

As shown in Table~\ref{tab:ddim_sampler}, our method with score scaling guidance (Ours$^{\text{DDIM+G}}$) demonstrates remarkable robustness across different sampling steps for single pose estimation. Even with only 10 sampling steps, we achieve 42.9 FPS while maintaining competitive accuracy. Notably, the performance remains nearly constant from 500 steps down to 50 steps, highlighting the effectiveness of our score scaling approach. In contrast, performance drops dramatically without scoring scaling guidance (Ours$^{\text{DDIM}}$: 52.7 → 15.9, 72.7 → 23.6),  further underscoring the critical role of our proposed guidance method.

Compared to recent forward-only baselines (HS-Pose~\cite{hspose} and Query6DoF~\cite{query6dof}), our approach offers a favorable speed-accuracy trade-off while using significantly fewer parameters, thus making it well-suited for real-time applications.

\begin{table}
\centering
\begin{adjustbox}{scale=0.65}
  \begin{tabular}{l | c | c c c c}
    \toprule
    Method  & Num steps  & Speed(FPS)$\uparrow$ & $5^\circ 2\text{cm}\uparrow$  & $10^\circ 2\text{cm}\uparrow$ &  Params(M)$\downarrow$\\
    \midrule
    Ours$^{\text{DDIM}}$  & 500  & 2.70 & 15.9 & 23.6  & 2.2\\
    \midrule
    Ours$^{\text{DDIM+G}}$  & 500  & 2.70 & 52.7 & 72.7  & 2.2\\
    Ours$^{\text{DDIM+G}}$  & 200  & 6.81 & 52.7 & 72.8  & 2.2\\
    Ours$^{\text{DDIM+G}}$  & 100  & 14.1 & 52.6 & 72.7  & 2.2\\
    Ours$^{\text{DDIM+G}}$  & 50  & 25.4 & 52.4 & 72.6  & 2.2\\
    Ours$^{\text{DDIM+G}}$  & 10  & 42.9 & 50.8 & 71.3  & 2.2\\
    \midrule
    HS-Pose~\cite{hspose}  & 1 & 50 & 46.5 & 68.6 & 6.1 \\
    Query6DoF~\cite{query6dof}  & 1 & 34.9 & 49.0 & 68.7 & 19.4\\
    \bottomrule
  \end{tabular}
\end{adjustbox}
\caption{Comparison of forward-only methods and ours with different sampling steps on REAL275 dataset for single frame pose estimation. For Ours, $^{\text{DDIM}}$ indicates DDIM Sampler without score scaling, and $^{\text{DDIM+G}}$ indicates DDIM Sampler with score scaling guidance.}
\label{tab:ddim_sampler}
\vspace{-0.3em}
\end{table}

\subsection{Analysis of Pre-training strategies}
Pre-training encoders has been widely adopted in various vision tasks to leverage learned representations, with notable success in models like Latent Diffusion Models (LDMs)~\cite{ldm} where pre-trained image encoders significantly reduce computational costs while maintaining generation quality. However, our investigation reveals that this established practice does not directly translate to category-level 6D pose estimation with point cloud inputs. While previous works like ~\cite{ssppose, gpvpose, rbppose, agpose, istnet} have utilized various point cloud encoders (e.g., 3DGC~\cite{3dgc}, PointNet++~\cite{pointnetpp}) trained in an end-to-end fashion, we systematically evaluated different pre-training strategies within the GenPose framework to potentially accelerate convergence and enhance performance.

As shown in Figure \ref{fig:four_plots}\subref{fig:a} in the main paper, we compared various pre-training strategies on NOCS dataset: classification of object categories (orange), point cloud reconstruction (green), and direct 6D pose regression (purple). Surprisingly, both classification and reconstruction pre-training show marginal or even negative impact on convergence compared to training from scratch (While Figure \ref{fig:four_plots}\subref{fig:a} shows results with PointNet++ encoder, similar patterns were observed when using Transformer-based encoder~\cite{pointmae}). In contrast, pre-training with direct 6D pose regression demonstrates notably faster convergence and better performance. We hypothesize that this phenomenon stems from the higher complexity of 6D pose estimation compared to classification or reconstruction tasks. Unlike classification which extracts category-discriminative features or reconstruction which preserves the local neighborhood structures~\cite{foldingnet}, 6D pose estimation demands the encoder to learn both fine-grained geometric features and their global spatial relationships in SE(3) space. When pre-trained on other tasks, the encoder learns features that may be suboptimal or even counterproductive for pose estimation. This observation led to our final design choice of joint learning with pose regression, which effectively combines the benefits of pre-trained pose-aware features with diffusion-based distribution modeling.



\begin{figure}[htp!]
  \centering
   \includegraphics[width=0.8\linewidth]{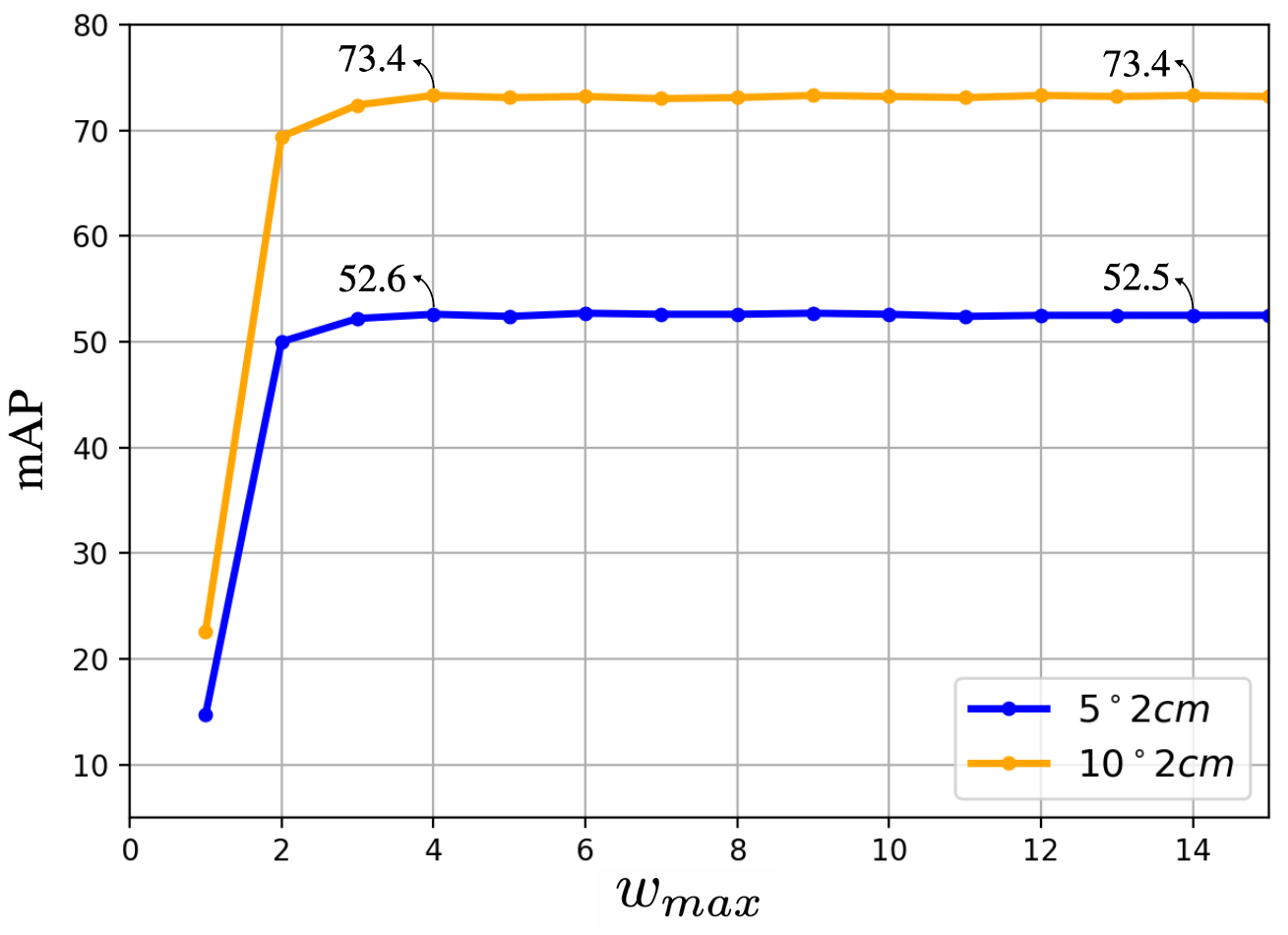}
   \caption{Performance on different weight parameters. All results are based on single pose sampling.}
   \label{fig:weight_ablation}
\end{figure}

\begin{figure*}[htp!]
     \centering
     \includegraphics[width=\textwidth]{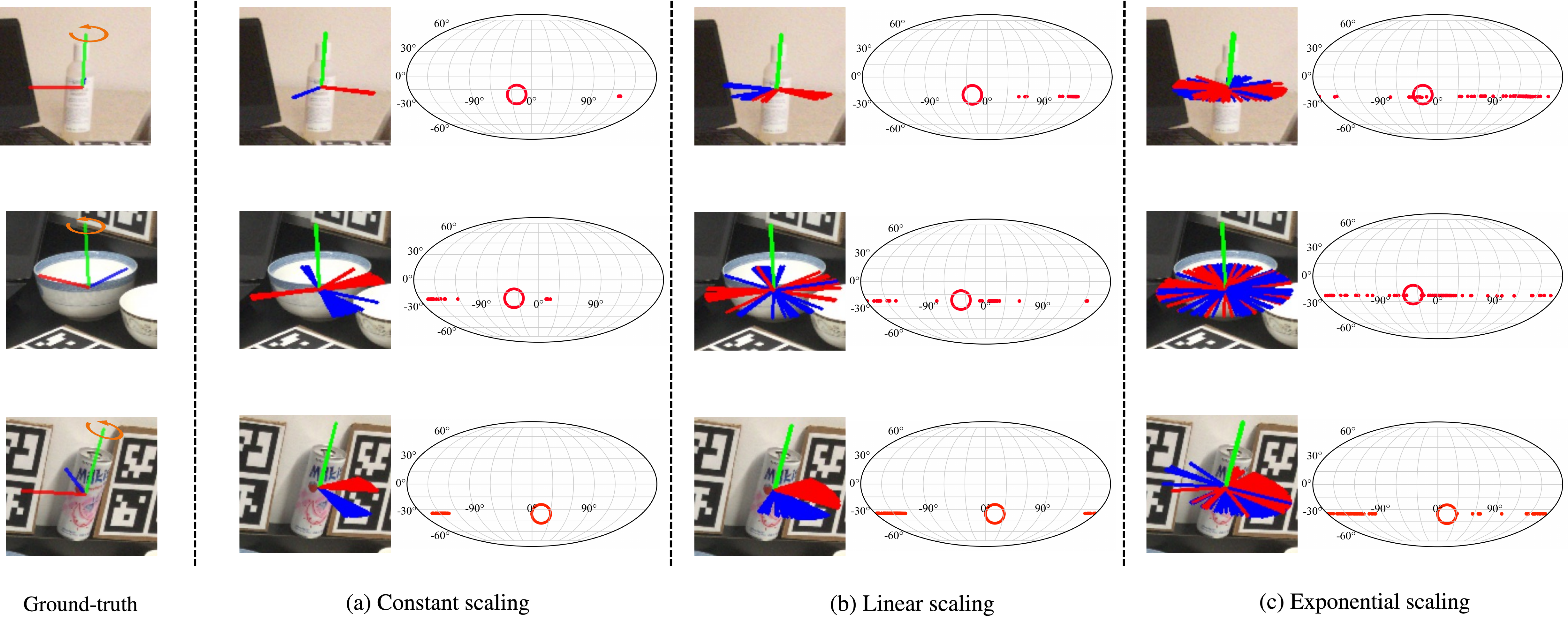}
     \caption{Comparison of rotation distributions for 50 sampled poses across different scaling schedules.}
     \label{fig:scheduler_compare}
 \end{figure*}

\subsection{Ablation studies on Scaling Guidance}

\paragraph{Comparison of Scaling Schedulers.}
To analyze the impact of guidance scheduling, we examine three different score scaling strategies:
\begin{equation} \label{eq:supp4}
w_t = \begin{cases} w_{max} + (w_{min}-w_{max}) t & (\text{linear}) \\ w_{min} + (w_{max}-w_{min})\exp(-5t) & (\text{exponential}) \\  w_{max} &  (\text{constant})
\end{cases}
\end{equation} 
Here, we set {\small\smash{$w_{min}=1.0$}} and {\small\smash{$w_{max}=4.0$}}. 
Table \ref{tab:scheduler_compare_table} shows similar quantitative performance across different schedulers on REAL275 dataset, though Figure \ref{fig:scheduler_compare} reveals their distinct behaviors when handling symmetric objects. Using 50 identical random noise inputs, the constant scheduler (a) leads to mode collapse with strong convergence to specific modes. The linear scheduler (b), which gradually increases guidance weight, shows improved but still limited capability in capturing the full symmetric distribution. In contrast, the exponential scheduler (c), which begins with weak guidance and gradually increases over time, best preserves the symmetric distribution while maintaining pose fidelity. This observation highlights that weak guidance in early time steps is crucial for exploring the symmetric distribution space, while stronger guidance near \smash{$t \rightarrow 0$} helps improve pose quality.

\begin{table}[htp!]
\begin{adjustbox}{scale=0.95}
  \centering
  \small
  \begin{tabular}{l | c c c c}
    \toprule
    Guidance Scheduler & $5^\circ 2\text{cm}$ & $5^\circ 5\text{cm}$ & $10^\circ 2\text{cm} $ & $10^\circ 5\text{cm}$\\
    \midrule
    Constant  &  {52.7} & {61.2} & {73.2}  & {84.5}\\
    Linear  & {52.7} & {61.1} & {73.2}  & {84.4}\\
    Exponential & {52.6} & {61.0} & {73.4}  & {84.5}\\
    \bottomrule
\end{tabular}
\end{adjustbox}
\caption{Results on REAL275 dataset according to score scaling weight scheduler.}
\label{tab:scheduler_compare_table}
\end{table}


\paragraph{Choice of the weight parameter for Guidance.}
We analyze the impact of the maximum weight parameter {\small\smash{$w_{max}$}} on model performance while fixing {\small\smash{$w_{min}=1$}} and using exponential scheduling. As shown in Figure \ref{fig:weight_ablation}, pose estimation accuracy improves as {\small\smash{$w_{max}$}} increases from 1 to 4 on the REAL275 dataset ({\small\smash{$w_{max}=1$}} refers to w/o Guidance). However, the performance plateaus beyond {\small\smash{$w_{max}=4$}}, with minimal or no improvements for larger values. We set {\small\smash{$w_{max}=4$}} as our default value based on these empirical results.

\subsection{Generalization ability}
Following GenPose~\cite{genpose}, we evaluate our model on unseen categories in REAL275, focusing on symmetric objects. As presented in Table~\ref{tab:generalization}, our method exhibits generalization capabilities comparable to those of GenPose, whereas other baseline~\cite{sarnet} suffers significant performance degradation. This indicates that diffusion-based models can generalize effectively to unseen object categories, particularly when they share geometric structures with the training set.

\begin{table}[htp!]
\centering
\begin{adjustbox}{scale=0.8}
  \small
  \begin{tabular}{c | c | c c c c}
    \toprule
    Category & Method & $5^\circ 2\text{cm}$ &$5^\circ 5\text{cm}$ & $10^\circ 2\text{cm}$  & $10^\circ 5\text{cm}$\\
    \midrule
    \multirow{3}{*}{bowl}  
    & SAR-Net & 58.1/36.4 & 66.0/47.3 & 83.7/59.4 & 93.6/81.5 \\
    & GenPose & 85.4/64.5 & 92.6/72.5 & 93.1/87.2 & 100.0/98.6 \\
    & Ours     & 87.4/65.7 & 93.4/72.3 & 93.9/88.4 & 100.0/99.1 \\
    \midrule
    \multirow{3}{*}{bottle}  
    & SAR-Net & 43.5/11.7 & 54.0/23.0 & 61.3/33.6 & 79.8/68.0 \\
    & GenPose & 52.6/39.0 & 60.9/53.2 & 81.4/73.6 & 92.7/94.6 \\
    & Ours     & 54.8/42.1 & 65.4/57.0 & 81.3/74.4 & 93.4/93.1 \\
    \bottomrule
\end{tabular}
\end{adjustbox}
\caption{Cross-category results on REAL275. Left and right of ‘/’ denote seen and unseen category performance, respectively.}
\label{tab:generalization}
\end{table}

\begin{figure*}[htp!]
     \centering
     \includegraphics[width=\textwidth]{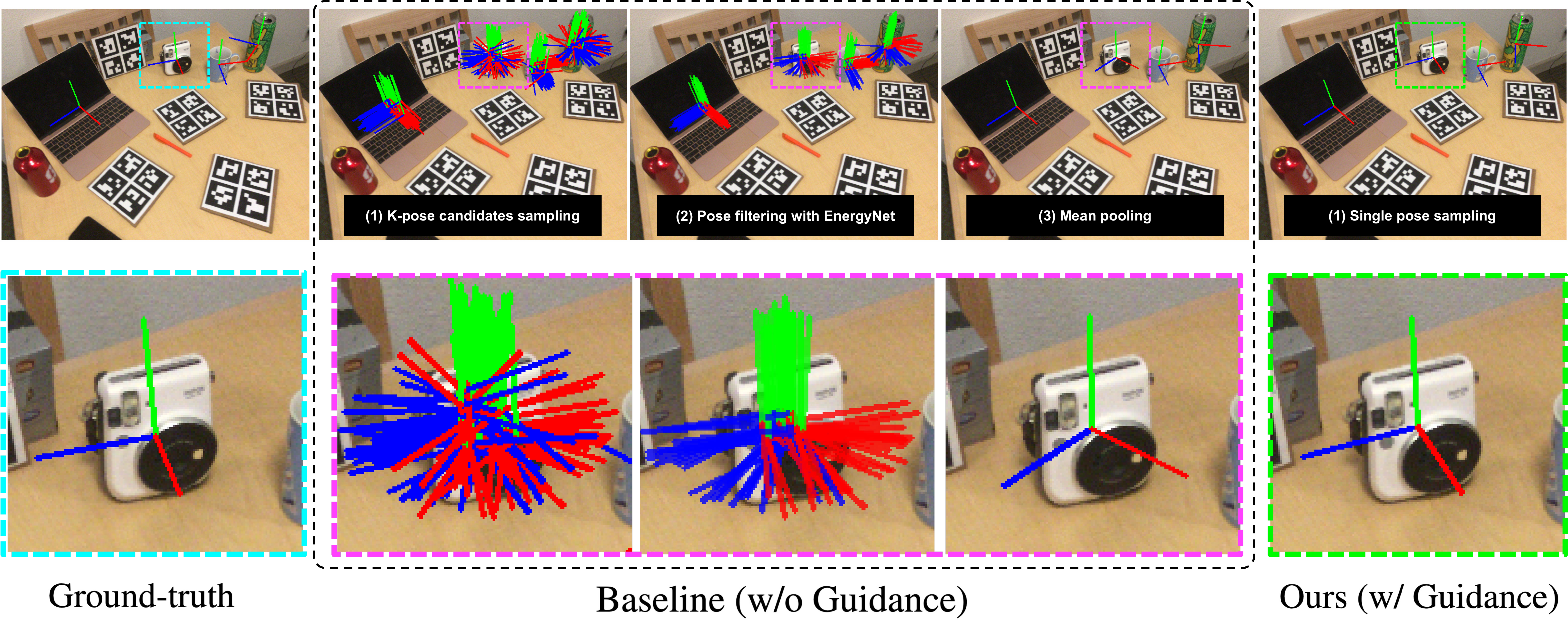}
     \caption{Qualitative comparison of Ours (column 5) and Baseline (column 2$\sim$4)}
     \label{fig:baseline_compare}
 \end{figure*}

\section{Additional Qualitative Results}
\label{sec:qual}
Figure \ref{fig:baseline_compare} provides a qualitative comparison between our method and the baseline (GenPose), illustrating their distinct sampling processes. The baseline's approach (shown in pink dashed box) requires multiple steps: (1) sampling K(=50) pose candidates, (2) filtering out low-likelihood poses using an additional EnergyNet, and (3) computing the final pose through mean pooling the remaining 30 poses. In contrast, our method (shown in green dashed box) employs score scaling guidance to generate high-quality poses with just a single pose sampling, eliminating the need for multiple pose candidates and additional filtering networks. 

\begin{figure*}[htp!]
    \vspace{-8.0em}
     \centering
     \includegraphics[width=\textwidth]{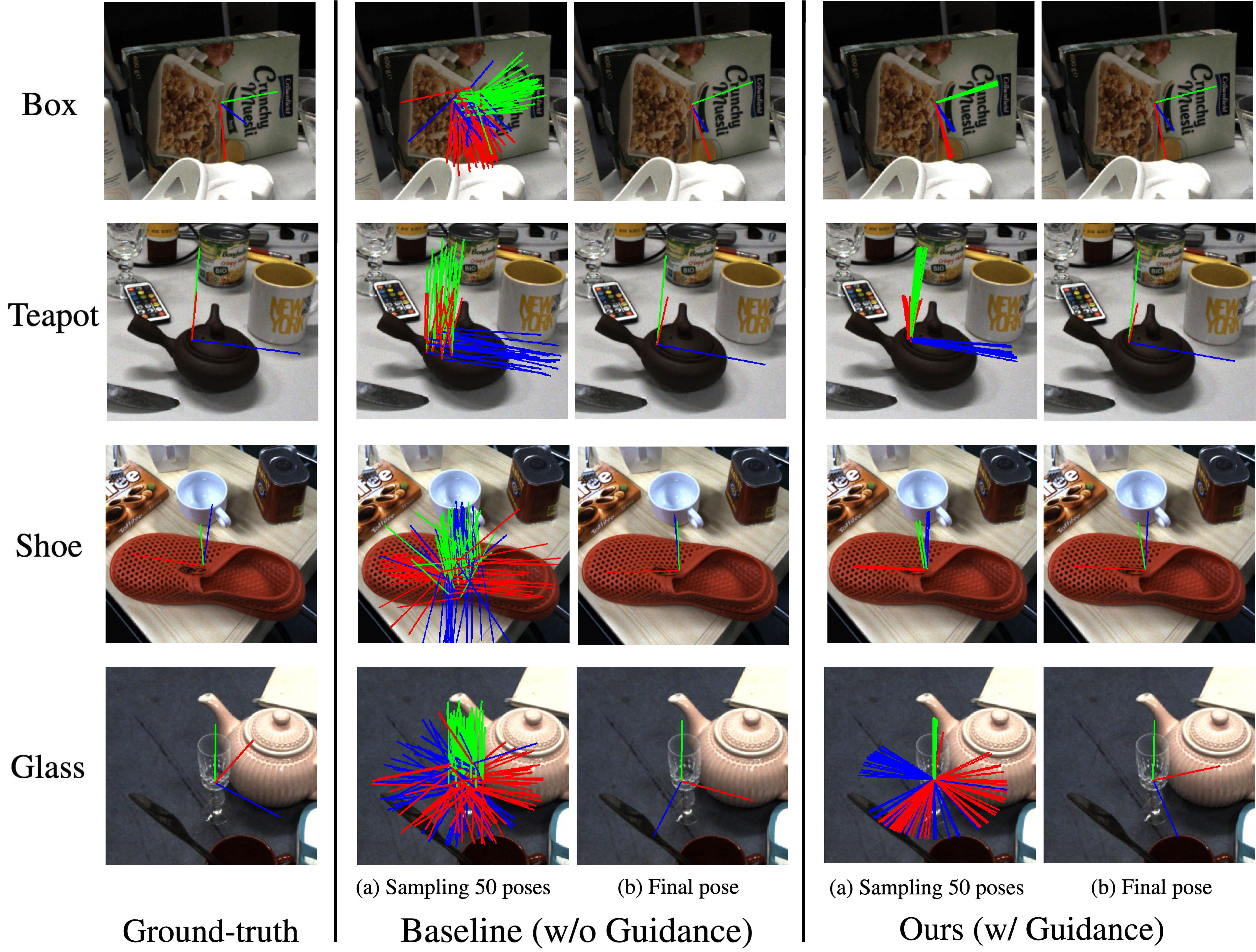}
     \caption{Qualitative comparison between Ours (column 3$\sim$4) and Baseline (column 1$\sim$2) on HouseCat6D.}
     \label{fig:qualitative_housecat6d}
 \end{figure*}

Figure \ref{fig:qualitative_housecat6d} presents additional qualitative comparisons between our method and GenPose on the HouseCat6D. GenPose (columns 1-2) first samples 50 pose candidates and computes the final output through filtering and mean pooling. While our method (columns 3-4) also samples 50 poses for comparison with the baseline, we randomly select a single pose to visualize the final output, demonstrating the effectiveness of our score scaling guidance.
For asymmetric objects (Box, Teapot, Shoe), while the baseline generates outlier samples that deviate from the ground truth pose (column 1), our method consistently produces pose samples that closely align with the ground truth (column 3). This improvement can be attributed to our guidance method, which effectively steers the sampling process toward high-density regions of the pose distribution.
The results for symmetric objects, particularly the Glass example, further highlight the advantages of our method. The baseline shows scattered pose samples around the symmetric axis, while our method with score scaling guidance accurately captures the object's symmetry. Specifically, our method precisely identifies the y-axis as the axis of symmetry (column 3), maintaining appropriate pose diversity while ensuring high-quality predictions.

\begin{figure*}[tp!]
    \vspace{-5.0em}
     \centering
     \includegraphics[width=\textwidth]{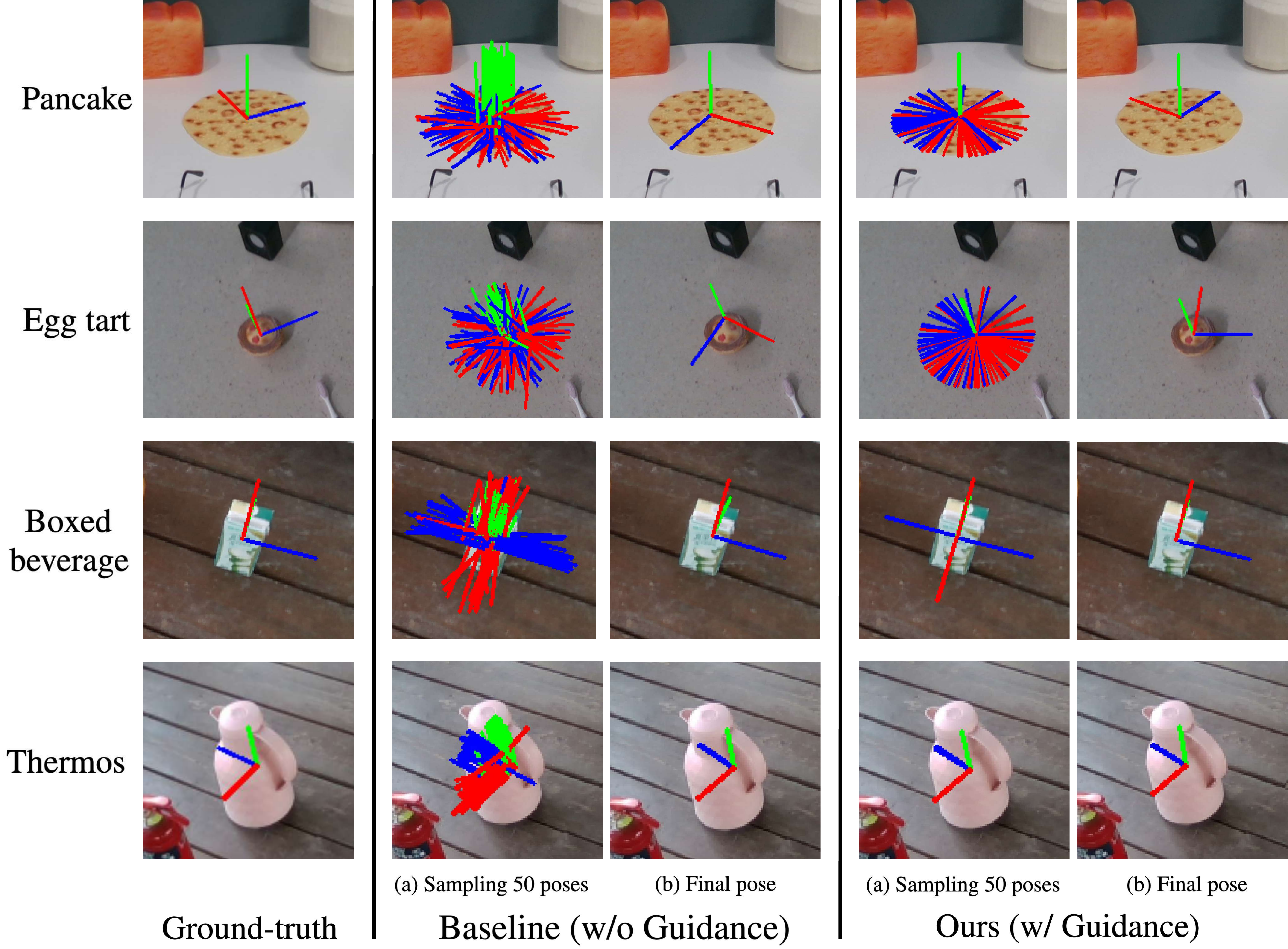}
     \caption{Qualitative comparison between Ours (column 3$\sim$4) and Baseline (column 1$\sim$2) on ROPE.}
     \label{fig:qualitative_rope}
 \end{figure*}

Figure \ref{fig:qualitative_rope} further demonstrates qualitative results on the ROPE dataset, highlighting our method's capability to handle objects with discrete symmetries. In the case of `Boxed beverage', which inherently has two symmetric ground truth poses (front and back), our method (column 3) successfully captures both valid pose modes. This demonstrates that our score scaling guidance effectively preserves the multi-modal nature of the pose distribution when dealing with objects that have multiple ground truth poses due to symmetry.


{
\small
~
}

\end{document}